\documentclass[12pt]{memoir}
\usepackage{amssymb}
\usepackage{amsmath}
\usepackage{amsthm}
\usepackage{url}
\usepackage{xspace}
\usepackage[margin=2.5cm]{geometry}
\usepackage{tikz}
\usetikzlibrary{positioning,calc,matrix,arrows}
\usepackage{pgfplots}
\usepackage{listings}
\usepackage{color}
\usepackage{textcomp}
\usepackage{fancyvrb}
\usepackage{array}
\usepackage[colorlinks]{hyperref}
\usepackage{cleveref}

\newtheorem{example}{Example}

\newcommand{\real}{\mathbb{R}}
\newcommand{\vv}{\operatorname{vec}}
\newcommand{\diag}{\operatorname{diag}}
\newcommand{\matconvnet}{\textsc{MatConvNet}\xspace}
\newcommand{\matlab}{\textsc{MATLAB}\xspace}

\newcommand{\bx}{\mathbf{x}}
\newcommand{\by}{\mathbf{y}}
\newcommand{\bc}{\mathbf{c}}

\newcommand{\bff}{\mathbf{f}}
\newcommand{\bg}{\mathbf{g}}

\newcommand{\bw}{\mathbf{w}}
\newcommand{\bp}{\mathbf{p}}
\newcommand{\bfs}{\mathbf{s}}

\newcommand{\bone}{\mathbf{1}}

\newcommand{\argmax}{\operatornamewithlimits{argmax}}
\newcommand{\sign}{\operatornamewithlimits{sign}}

\tikzstyle{block} = [draw, rectangle, minimum height=3em, minimum width=3em]
\tikzstyle{data} = []
\tikzstyle{datac} = [draw, circle, minimum height=2.5em, minimum width=2.5em,inner sep=3pt,font=\footnotesize]
\tikzstyle{par} = [draw, circle, minimum height=2.5em, minimum width=2.5em,fill=black!20,inner sep=3pt,font=\footnotesize]
\tikzstyle{pinstyle} = [pin edge={to-,thin,black}]
\tikzstyle{to} = [->,>=stealth',shorten >=1pt,semithick]
\tikzstyle{from} = [<-,>=stealth',shorten >=1pt,semithick]
\tikzstyle{bp} = [draw=blue,text=blue]
\tikzstyle{bpl} = [draw=blue!40]
\tikzstyle{bpe} = [text=blue,draw=none]

\VerbatimFootnotes

\setsecnumdepth{subsection}
\settocdepth{subsection}

\definecolor{listinggray}{gray}{0.9}
\definecolor{lbcolor}{rgb}{0.8,0.8,0.8}
\lstMakeShortInline[columns=fullflexible, breaklines = true,
 breakatwhitespace = true]!

\title{MatConvNet \\
\Large
Convolutional Neural Networks for MATLAB}
\author{
Andrea Vedaldi
\and
Karel Lenc}
\date{}

\hypersetup{
  pdfinfo={
    Title={MatConvNet: Convolutional Neural Networks for MATLAB},
    Author={Andrea Vedaldi and Karel Lenc},
		Subject={Reference Manual},
    Keywords={computer vision, convolutional neural networks, deep learning, image understanding, machine learning}
  }
}

\begin{document}


\frontmatter
\maketitle{}
\clearpage

\begin{abstract}
\matconvnet is an implementation of Convolutional Neural Networks (CNNs) for MATLAB. The toolbox is designed with an emphasis on simplicity and flexibility. It exposes the building blocks of CNNs as easy-to-use MATLAB functions, providing routines for computing linear convolutions with filter banks, feature pooling, and many more. In this manner, \matconvnet allows fast prototyping of new CNN architectures; at the same time, it supports efficient computation on CPU and GPU allowing to train complex models on large datasets such as ImageNet ILSVRC. This document provides an overview of CNNs and how they are implemented in \matconvnet and gives the technical details of each computational block in the toolbox.
\end{abstract}
\clearpage

\tableofcontents*
\clearpage

\mainmatter
\chapter{Introduction to MatConvNet}\label{s:intro}

\matconvnet is a MATLAB toolbox implementing \emph{Convolutional Neural Networks} (CNN) for computer vision applications.  Since the breakthrough work of~\cite{krizhevsky12imagenet}, CNNs have had a major impact in computer vision, and image understanding in particular, essentially replacing traditional image representations such as the ones implemented in our own VLFeat~\cite{vedaldi10vlfeat} open source library.

While most CNNs are  obtained by composing simple linear and non-linear filtering operations such as convolution and rectification, their implementation is far from trivial. The reason is that CNNs need to be learned from vast amounts of data, often millions of images, requiring very efficient implementations. As most CNN libraries, \matconvnet achieves this by using a variety of optimizations and, chiefly, by supporting computations on GPUs.

Numerous other machine learning, deep learning, and CNN open source libraries exist. To cite some of the most popular ones: CudaConvNet,\footnote{\small\url{https://code.google.com/p/cuda-convnet/ }} Torch,\footnote{\small\url{http://cilvr.nyu.edu/doku.php?id=code:start}} Theano,\footnote{\small\url{http://deeplearning.net/software/theano/}} and Caffe\footnote{\small\url{http://caffe.berkeleyvision.org}}.  Many of these libraries are  well supported, with dozens of active contributors and large user bases. Therefore, why creating yet another library?

The key motivation for developing \matconvnet was to provide an environment particularly friendly and efficient for researchers to use in their investigations.\footnote{While from a user perspective \matconvnet currently relies on MATLAB, the library is being developed with a clean separation between MATLAB code and the C++ and CUDA core; therefore, in the future the library may be extended to allow processing convolutional networks independently of MATLAB.} \matconvnet achieves this by its deep integration in the MATLAB environment, which is one of the most popular development environments in computer vision research as well as in many other areas. In particular, \matconvnet exposes as simple MATLAB commands CNN building blocks such as convolution, normalisation and pooling (\cref{s:blocks}); these can then be combined and extended with ease to create CNN architectures. While many of such blocks use optimised CPU and GPU implementations written in C++ and CUDA (section~\cref{s:speed}), MATLAB native support for GPU computation means that it is often possible to write new blocks in MATLAB directly while maintaining computational efficiency. Compared to writing new CNN components using lower level languages, this is an important simplification that can significantly accelerate testing new ideas. Using MATLAB also provides a bridge towards other areas; for instance, \matconvnet was recently used by the University of Arizona in planetary science, as summarised in this NVIDIA blogpost.\footnote{\small\url{http://devblogs.nvidia.com/parallelforall/deep-learning-image-understanding-planetary-science/}}

\matconvnet can learn large CNN models such AlexNet~\cite{krizhevsky12imagenet} and the very deep networks of~\cite{simonyan14deep} from millions of images. Pre-trained versions of several of these powerful models can be downloaded from  the \matconvnet home page\footnote{\small\url{http://www.vlfeat.org/matconvnet/}}. While powerful, \matconvnet remains simple to use and install. The implementation is fully self-contained, requiring only MATLAB and a compatible C++ compiler (using the GPU code requires the freely-available CUDA DevKit and a suitable NVIDIA GPU). As demonstrated in \cref{f:demo} and \cref{s:getting-statrted}, it is possible to download, compile, and install \matconvnet using three MATLAB commands. Several fully-functional examples demonstrating how small and large networks can be learned are included. Importantly, several \emph{standard pre-trained network} can be immediately downloaded and used in applications. A manual with a complete technical description of the toolbox is maintained along with the toolbox.\footnote{\small\url{http://www.vlfeat.org/matconvnet/matconvnet-manual.pdf}} These features make \matconvnet useful in an educational context too.\footnote{An example laboratory experience based on \matconvnet can be downloaded from {\small\url{http://www.robots.ox.ac.uk/~vgg/practicals/cnn/index.html}}.}

\matconvnet is open-source released under a BSD-like license. It can be downloaded from \url{http://www.vlfeat.org/matconvnet} as well as from GitHub.\footnote{\small\url{http://www.github.com/matconvnet}}.

\section{Getting started}\label{s:getting-statrted}

\begin{figure}
\hrule
\begin{lstlisting}[escapechar=!]
% install and compile MatConvNet (run once)
untar(['http://www.vlfeat.org/matconvnet/download/' ...
   'matconvnet-1.0-beta12.tar.gz']) ;
cd matconvnet-1.0-beta12
run matlab/vl_compilenn

% download a pre-trained CNN from the web (run once)
urlwrite(...
 'http://www.vlfeat.org/matconvnet/models/imagenet-vgg-f.mat', ...
 'imagenet-vgg-f.mat') ;

% setup MatConvNet
run matlab/vl_setupnn

% load the pre-trained CNN
net = load('imagenet-vgg-f.mat') ;

% load and preprocess an image
im = imread('peppers.png') ;
im_ = imresize(single(im), net.meta.normalization.imageSize(1:2)) ;
im_ = im_ - net.meta.normalization.averageImage ;

% run the CNN
res = vl_simplenn(net, im_) ;

% show the classification result
scores = squeeze(gather(res(end).x)) ;
[bestScore, best] = max(scores) ;
figure(1) ; clf ; imagesc(im) ;!
\begin{tikzpicture}[overlay]
\node (x) {};
\node (y) at (7,1) {\includegraphics[width=5cm]{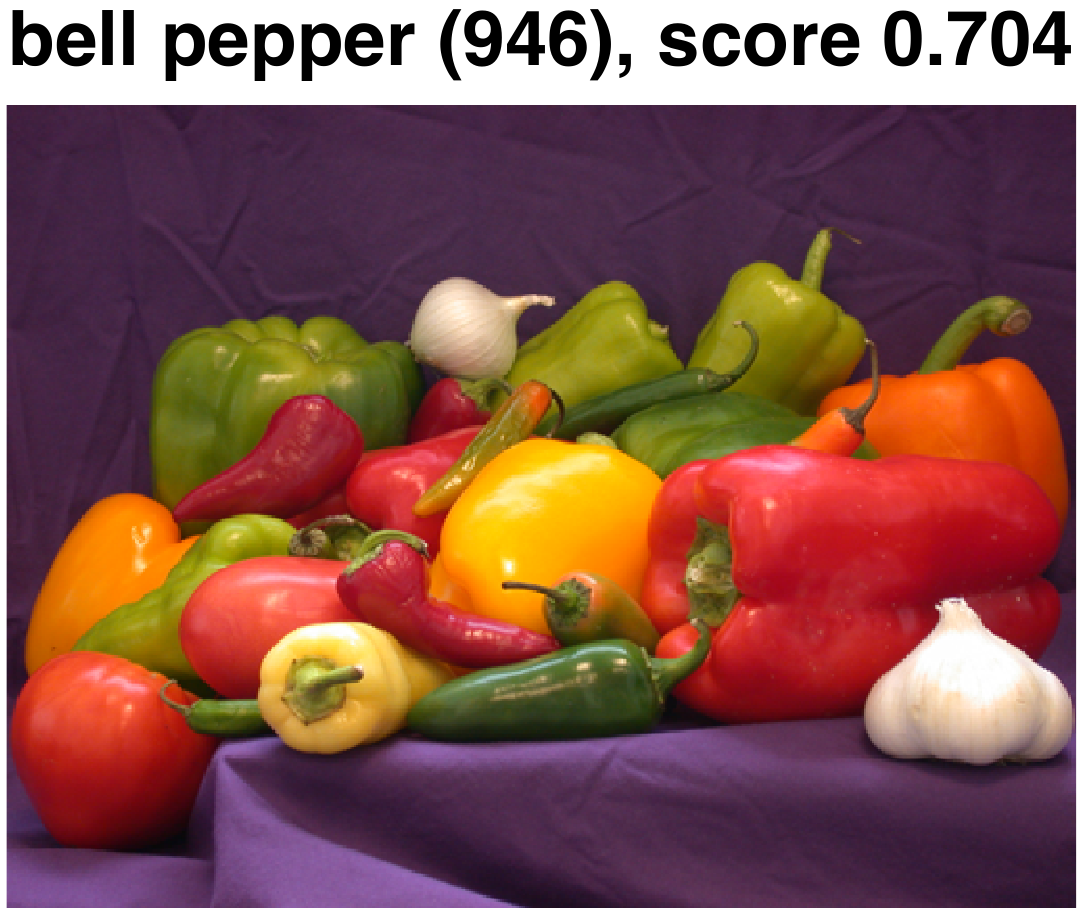}};
\draw [->,thick] (.1,.1) -- (4.5,.1) {};
\end{tikzpicture}!
title(sprintf('%s (%d), score %.3f',...
net.classes.description{best}, best, bestScore)) ;
\end{lstlisting}
\hrule
\caption{A complete example including download, installing, compiling and running  \matconvnet to classify one of  \matlab stock images using a large CNN pre-trained on ImageNet.}
\label{f:demo}
\end{figure}

\matconvnet is simple to install and use. \cref{f:demo} provides a complete example that classifies an image using a latest-generation deep convolutional neural network. The example includes downloading MatConvNet, compiling the package, downloading a pre-trained CNN model, and evaluating the latter on one of \matlab's stock images.

The key command in this example is !vl_simplenn!, a wrapper that takes as input the CNN !net! and the pre-processed image !im_! and produces as output a structure !res! of results. This particular wrapper can be used to model networks that have a simple structure, namely a \emph{chain} of operations. Examining the code of !vl_simplenn! (!edit vl_simplenn! in \matconvnet) we note that the wrapper transforms the data sequentially, applying a number of \matlab functions as specified by the network configuration. These function, discussed in detail in \cref{s:blocks}, are called ``building blocks'' and constitute the backbone of \matconvnet.


While most blocks implement simple operations, what makes them non trivial is their efficiency (\cref{s:speed}) as well as support for backpropagation (\cref{s:back}) to allow learning CNNs. Next, we demonstrate how to use one of such building blocks directly. For the sake of the example, consider convolving an image with a bank of linear filters. Start by reading an image in \matlab, say using !im = single(imread('peppers.png'))!, obtaining a $H \times W \times D$ array !im!, where $D=3$ is the number of colour channels in the image. Then create a bank of $K=16$ random filters of size $3 \times 3$ using !f = randn(3,3,3,16,'single')!. Finally, convolve the image with the filters by using the command !y = vl_nnconv(x,f,[])!. This results in an array !y! with $K$ channels, one for each of the $K$ filters in the bank.

While users are encouraged to make use of the blocks directly to create new architectures, \matlab provides wrappers such as !vl_simplenn! for standard CNN architectures such as AlexNet~\cite{krizhevsky12imagenet} or Network-in-Network~\cite{lin13network}. Furthermore, the library provides numerous examples (in the !examples/! subdirectory), including code to learn a variety of models on the MNIST, CIFAR, and ImageNet datasets. All these examples use the !examples/cnn_train! training code, which is an implementation of stochastic gradient descent (\cref{s:wrappers-learning}). While this training code is perfectly serviceable and quite flexible, it remains in the !examples/! subdirectory as it is somewhat problem-specific. Users are welcome to implement their optimisers.

\section{\matconvnet at a glance}\label{s:vlnn}

\matconvnet has a simple design philosophy. Rather than wrapping CNNs around complex layers of software, it exposes simple functions to compute CNN building blocks, such as linear convolution and ReLU operators, directly as MATLAB commands. These building blocks are easy to combine into complete CNNs and can be used to implement sophisticated learning algorithms. While several real-world examples of small and large CNN architectures and training routines are provided, it is always possible to go back to the basics and build your own, using the efficiency of MATLAB in prototyping. Often no C coding is required at all to try new architectures. As such, \matconvnet is an ideal playground for research in computer vision and CNNs.

\matconvnet contains the following elements:
\begin{itemize}
\item \emph{CNN computational blocks.} A set of optimized routines computing fundamental building blocks of a CNN. For example, a convolution block is implemented by \linebreak !y=vl_nnconv(x,f,b)! where !x! is an image, !f! a filter bank, and !b! a vector of biases (\cref{s:convolution}). The derivatives are computed as
![dzdx,dzdf,dzdb] = vl_nnconv(x,f,b,dzdy)! where !dzdy! is the derivative of the CNN output w.r.t !y!~(\cref{s:convolution}). \cref{s:blocks} describes all the blocks in detail.
\item \emph{CNN wrappers.} \matconvnet provides a simple wrapper, suitably invoked by !vl_simplenn!, that implements a CNN with a linear topology (a chain of blocks). It also provides a much more flexible wrapper supporting networks with arbitrary topologies, encapsulated in the !dagnn.DagNN! MATLAB class.
\item \emph{Example applications.} \matconvnet provides several examples of learning CNNs with stochastic gradient descent and CPU or GPU, on MNIST, CIFAR10, and ImageNet data.
\item \emph{Pre-trained models.} \matconvnet provides several state-of-the-art pre-trained CNN models that can be used off-the-shelf, either to classify images or to produce image encodings in the spirit of Caffe or DeCAF.
\end{itemize}

\section{Documentation and examples}\label{s:examples}

\begin{figure}
\centering
\includegraphics[width=0.65\columnwidth]{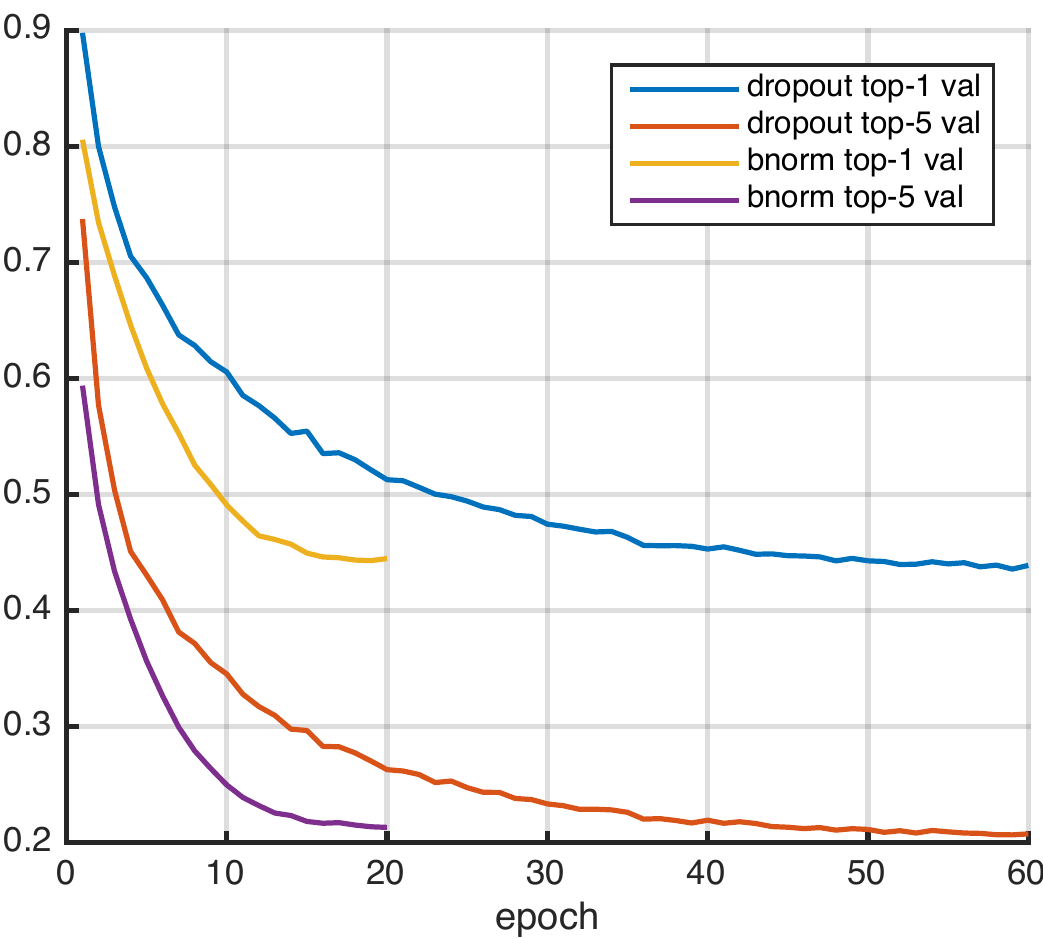}
\vspace{-1em}
\caption{Training AlexNet on ImageNet ILSVRC: dropout vs batch normalisation.}\label{f:imnet}
\end{figure}

There are three main sources of information about \matconvnet. First, the website contains descriptions of all the functions and several examples and tutorials.\footnote{\small See also \url{http://www.robots.ox.ac.uk/~vgg/practicals/cnn/index.html}.} Second, there is a PDF manual containing a great deal of technical details about the toolbox, including detailed mathematical descriptions of the building blocks. Third, \matconvnet ships with several examples (\cref{s:getting-statrted}).

Most examples are fully self-contained. For example, in order to run the MNIST example, it suffices to point MATLAB to the \matconvnet root directory and type !addpath examples! followed by !cnn_mnist!. Due to the problem size, the ImageNet ILSVRC example requires some more preparation, including downloading and preprocessing the images (using the bundled script !utils/preprocess-imagenet.sh!). Several advanced examples are included as well. For example, \cref{f:imnet} illustrates the top-1 and top-5 validation errors as a model similar to AlexNet~\cite{krizhevsky12imagenet} is trained using either standard dropout regularisation or the recent \emph{batch normalisation} technique of~\cite{ioffe15batch}. The latter is shown to converge in about one third of the epochs (passes through the training data) required by the former.

The \matconvnet website contains also numerous \emph{pre-trained} models, i.e. large CNNs trained on ImageNet ILSVRC that can be downloaded and used as a starting point for many other problems~\cite{chatfield14return}. These include: AlexNet~\cite{krizhevsky12imagenet}, VGG-S, VGG-M,  VGG-S~\cite{chatfield14return}, and  VGG-VD-16, and VGG-VD-19~\cite{simonyan15very}.  The example code of \cref{f:demo} shows how one such model can be used in a few lines of MATLAB code.

\section{Speed}\label{s:speed}

Efficiency is very important for working with CNNs. \matconvnet supports  using NVIDIA GPUs as it includes CUDA implementations of all algorithms (or relies on MATLAB CUDA support).

 To use the GPU (provided that suitable hardware is available and the toolbox has been compiled with GPU support), one simply converts the arguments to !gpuArrays! in MATLAB, as in !y = vl_nnconv(gpuArray(x), gpuArray(w), [])!. In this manner, switching between CPU and GPU is fully transparent. Note that \matconvnet can also make use of the NVIDIA CuDNN library with significant speed and space benefits.

Next we evaluate the performance of \matconvnet when training large architectures on the ImageNet ILSVRC 2012 challenge data~\cite{deng09imagenet}. The test machine is a Dell server with two Intel Xeon CPU E5-2667 v2 clocked at 3.30 GHz (each CPU has eight cores), 256 GB of RAM, and four NVIDIA Titan Black GPUs (only one of which is used unless otherwise noted). Experiments use \matconvnet beta12, CuDNN v2, and MATLAB R2015a. The data is preprocessed to avoid rescaling images on the fly in MATLAB and stored in a RAM disk for faster access. The code uses the !vl_imreadjpeg! command to read large batches of JPEG images from disk in a number of separate threads. The driver !examples/cnn_imagenet.m! is used in all experiments.

 We train the models discussed in \cref{s:examples} on ImageNet ILSVRC. \cref{f:speed} reports the training speed as number of images per second processed by stochastic gradient descent. AlexNet trains at about 264 images/s with CuDNN, which is about 40\% faster than the vanilla GPU implementation (using CuBLAS) and more than 10 times faster than using the CPUs. Furthermore, we note that, despite MATLAB overhead, the implementation speed is comparable to Caffe (they report 253 images/s with CuDNN and a Titan -- a slightly slower GPU than the Titan Black used here).  Note also that, as the model grows in size, the size of a SGD batch must be decreased (to fit in the GPU memory), increasing the overhead impact somewhat.

 \cref{f:mgpu} reports the speed on VGG-VD-16, a very large model, using multiple GPUs. In this case, the batch size is set to 264 images. These are further divided in sub-batches of 22 images each to fit in the GPU memory; the latter are then distributed among one to four GPUs on the same machine. While there is a substantial communication overhead, training speed increases from 20 images/s to 45. Addressing this overhead is one of the medium term goals of the library.

\begin{table}
\centering
\begin{tabular}{|lc|ccc|}
  \hline
  model     & batch sz. & CPU  & GPU   & CuDNN \\
  \hline
  AlexNet   & 256       & 22.1 & 192.4 & 264.1 \\
  VGG-F     & 256       & 21.4 & 211.4 & 289.7 \\
  VGG-M     & 128       & 7.8  & 116.5 & 136.6 \\
  VGG-S     & 128       & 7.4  & 96.2  & 110.1 \\
  VGG-VD-16 & 24        & 1.7  & 18.4  & 20.0  \\
  VGG-VD-19 & 24        & 1.5  & 15.7  & 16.5  \\
  \hline
\end{tabular}
\caption{ImageNet training speed (images/s).}
\label{f:speed}
\end{table}

\begin{table}
\centering
\begin{tabular}{|c|cccc|}
  \hline
  num GPUs     & 1  & 2 & 3 & 4 \\
  \hline
  VGG-VD-16 speed & 20.0 & 22.20 & 38.18 & 44.8 \\
  \hline
\end{tabular}
\caption{Multiple GPU speed (images/s).}
\label{f:mgpu}
\end{table}

\section{Acknowledgments}\label{s:ack}

\matconvnet is a community project, and as such acknowledgements go to all contributors. We kindly thank NVIDIA supporting this project by providing us with top-of-the-line GPUs and MathWorks for ongoing discussion on how to improve the library.

The implementation of several CNN computations in this library are inspired by the Caffe library~\cite{jia13caffe} (however, Caffe is \emph{not} a dependency). Several of the example networks have been trained by Karen Simonyan as part of~\cite{chatfield14return} and~\cite{simonyan15very}.

\chapter{Neural Network Computations}\label{s:fundamentals}

This chapter provides a brief introduction to the computational aspects of neural networks, and convolutional neural networks in particular, emphasizing the concepts required to understand and use \matconvnet.

\section{Overview}\label{s:cnn-structure}

A \emph{Neural Network} (NN) is a function $g$ mapping data $\bx$, for example an image, to an output vector $\by$, for example an image label. The function $g=f_L \circ \dots \circ f_1$ is the composition of a sequence of simpler functions $f_l$, which are called \emph{computational blocks} or \emph{layers}. Let $\bx_1,\bx_2,\dots,\bx_L$ be the outputs of each layer in the network, and let $\bx_0=\bx$ denote the network input. Each intermediate output $\bx_l = f_l(\bx_{l-1};\bw_l)$ is computed from the previous output $\bx_{l-1}$  by applying the function $f_l$ with parameters $\bw_l$. 

In a \emph{Convolutional Neural Network} (CNN), the data has a spatial structure: each $\bx_l\in\mathbb{R}^{H_l \times W_l \times C_l}$ is a 3D array or \emph{tensor} where the first two dimensions $H_l$ (height) and $W_l$ (width) are interpreted as spatial dimensions. The third dimension $C_l$ is instead interpreted as the \emph{number of feature channels}. Hence, the tensor $\bx_l$ represents a $H_l \times W_l$ field of $C_l$-dimensional feature vectors, one for each spatial location. A fourth dimension $N_l$ in the tensor spans multiple data samples packed in a single \emph{batch} for efficiency parallel processing. The number of data samples $N_l$ in a batch is called the batch \emph{cardinality}. The network is called \emph{convolutional} because the functions $f_l$ are local and translation invariant operators (i.e.\ non-linear filters) like linear convolution.

It is also possible to conceive CNNs with more than two spatial dimensions, where the additional dimensions may represent volume or time. In fact, there are little \emph{a-priori} restrictions on the format of data in neural networks in general. Many useful NNs contain a mixture of convolutional layers together with layer that process other data types such as text strings, or perform other operations that do not strictly conform  to the CNN assumptions.

\matconvnet includes a variety of layers, contained in the !matlab/! directory, such as !vl_nnconv! (convolution), !vl_nnconvt! (convolution transpose or deconvolution), !vl_nnpool! (max and average pooling), !vl_nnrelu! (ReLU activation), !vl_nnsigmoid! (sigmoid activation), !vl_nnsoftmax! (softmax operator), !vl_nnloss! (classification log-loss), !vl_nnbnorm! (batch normalization), !vl_nnspnorm! (spatial normalization), !vl_nnnormalize! (locar response normalization -- LRN), or !vl_nnpdist! ($p$-distance).  There are enough layers to implement many interesting state-of-the-art networks out of the box, or even import them from other toolboxes such as Caffe. 

NNs are often used as classifiers or regressors. In the example of \cref{f:demo}, the output $\hat \by = f(\bx)$ is a vector of probabilities, one for each of a 1,000 possible image labels (dog, cat, trilobite, ...).  If $\by$ is the true label of image $\bx$, we can measure the CNN performance by a loss function $\ell_\by(\hat \by)  \in \mathbb{R}$ which assigns a penalty to classification errors. The CNN parameters can then be tuned or \emph{learned} to minimize this loss averaged over a large dataset of labelled example images.

Learning generally uses a variant of \emph{stochastic gradient descent} (SGD). While this is an efficient method (for this type of problems), networks may contain several million parameters and need to be trained on millions of images; thus, efficiency is a paramount in \matlab design, as further discussed in \cref{s:speed}. SGD also requires to compute the CNN derivatives, as explained in the next section.

\section{Network structures}\label{s:cnn-topology}

In the simplest case, layers in a NN are arranged in a sequence; however, more complex interconnections are possible as well, and in fact very useful in many cases. This section discusses such configurations and introduces a graphical notation to visualize them.

\subsection{Sequences}\label{s:cnn-simple}

Start by considering a computational block $f$ in the network. This can be represented schematically as a box receiving data $\bx$ and parameters $\bw$ as inputs and producing data $\by$ as output:
\begin{center}
\begin{tikzpicture}[auto, node distance=2cm]
\node (x) [data] {$\bx$};
\node (f) [block,right of=x]{$f$};
\node (y) [data, right of=f] {$\by$};
\node (w) [data, below of=f] {$\bw$};
\draw [to] (x.east) -- (f.west) {};
\draw [to] (f.east) -- (y.west) {};
\draw [to] (w.north) -- (f.south) {};
\end{tikzpicture}
\end{center}
As seen above, in the simplest case blocks are chained in a sequence $f_1 \rightarrow f_2\rightarrow\dots\rightarrow f_L$ yielding the structure:
\begin{center}
\begin{tikzpicture}[auto, node distance=2cm]
\node (x0)  [data] {$\bx_0$};
\node (f1) [block,right of=x0]{$f_1$};
\node (f2) [block,right of=f1,node distance=3cm]{$f_2$};
\node (dots) [right of=f2]{...};
\node (fL) [block,right of=dots]{$f_L$};
\node (xL)  [data, right of=fL] {$\bx_L$};
\node (w1) [data, below of=f1] {$\bw_1$};
\node (w2) [data, below of=f2] {$\bw_2$};
\node (wL) [data, below of=fL] {$\bw_L$};
\draw [to] (x0.east) -- (f1.west) {};
\draw [to] (f1.east) -- node {$\bx_1$} (f2.west);
\draw [to] (f2.east) -- node {$\bx_2$} (dots.west) {};
\draw [to] (dots.east) -- node {$\bx_{L-1}$} (fL.west) {};
\draw [to] (fL.east) -- (xL.west) {};
\draw [to] (w1.north) -- (f1.south) {};
\draw [to] (w2.north) -- (f2.south) {};
\draw [to] (wL.north) -- (fL.south) {};
\end{tikzpicture}
\end{center}
Given an input $\bx_0$, evaluating the network is a simple matter of evaluating all the blocks from left to right, which defines a composite function $\bx_L = f(\bx_0;\bw_1,\dots,\bw_L)$. 

\subsection{Directed acyclic graphs}\label{s:cnn-dag}

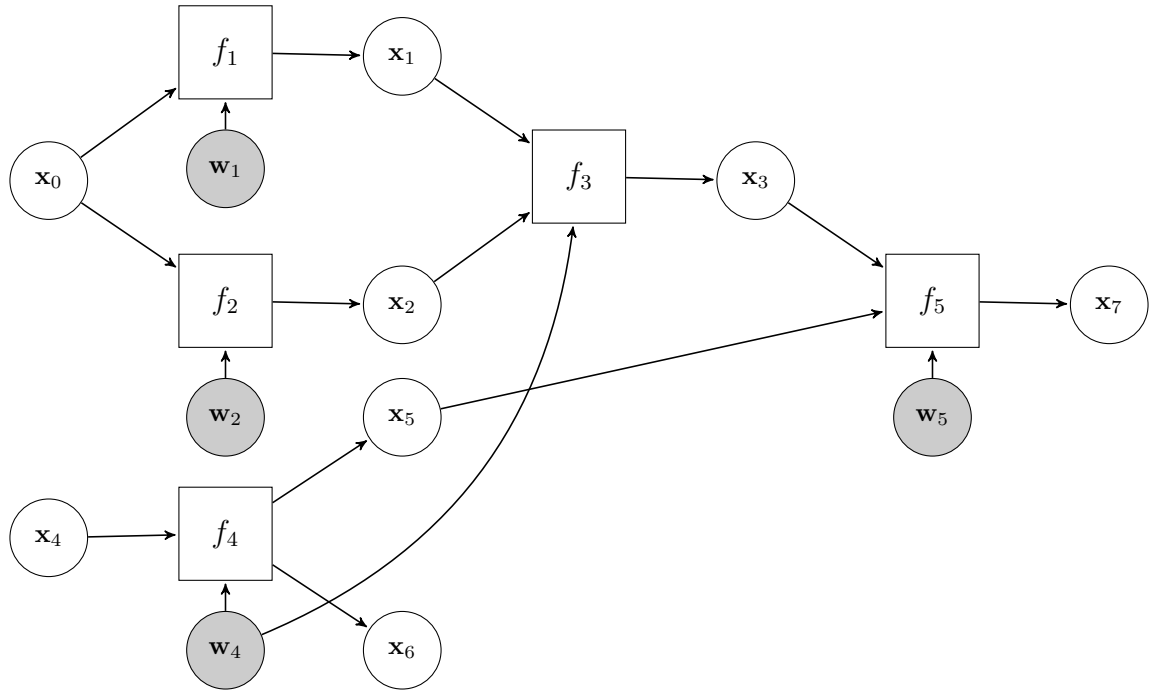
\begin{figure}[t]
\begin{center}
\begin{tikzpicture}[auto, node distance=0.4cm]
 \matrix (m) [matrix of math nodes, 
    column sep=1.2cm,
    row sep=0.4cm]
{
& \node (f1) [block]{f_1}; 
& \node (x1) [datac]{\bx_1};
\\
\node (x0) [datac]{\bx_0};
&
&
& \node (f3) [block]{f_3};
& \node (x3) [datac]{\bx_3};
\\
& \node (f2) [block]{f_2}; 
& \node (x2) [datac]{\bx_2};
& &
& \node (f5) [block]{f_5}; 
& \node (x7) [datac]{\bx_7}; 
\\
& 
& \node(x5) [datac]{\bx_5};
\\
\node (x4) [datac]{\bx_4};
& \node (f4) [block]{f_4};
\\
& 
& \node(x6) [datac]{\bx_6};
\\
};
\draw[to] (x0) -- (f1);
\draw[to] (f1) -- (x1);
\draw[to] (x1) -- (f3);
\draw[to] (x0) -- (f2);
\draw[to] (f2) -- (x2);
\draw[to] (x2) -- (f3);
\draw[to] (f3) -- (x3);
\draw[to] (x3) -- (f5);
\draw[to] (f5) -- (x7);
\draw[to] (x4) -- (f4);
\draw[to] (f4) -- (x5);
\draw[to] (f4) -- (x6);
\draw[to] (x5) -- (f5);
\node(w1) [par,below=of f1]{$\bw_1$}; \draw[to] (w1) -- (f1);
\node(w2) [par,below=of f2]{$\bw_2$}; \draw[to] (w2) -- (f2);
\node(w4) [par,below=of f4]{$\bw_4$}; \draw[to] (w4) -- (f4);
\draw[to] (w4) to [bend right] (f3);
\node(w5) [par,below=of f5]{$\bw_5$}; \draw[to] (w5) -- (f5);
\end{tikzpicture}
\end{center}
\vspace{-1em}
\caption{\textbf{Example DAG.}}\label{f:dag}
\end{figure}

One is not limited to chaining layers one after another. In fact, the only requirement for evaluating a NN is that, when a layer has to be evaluated, all its input have been evaluated prior to it. This is possible exactly when the interconnections between layers form a \emph{directed acyclic graph}, or DAG for short.

In order to visualize DAGs, it is useful to introduce additional nodes for the network variables, as in the  example of Fig.~\ref{f:dag}. Here boxes denote functions and circles denote variables (parameters are treated as a special kind of variables). In the example, $\bx_0$ and $\bx_4$ are the inputs of the CNN and $\bx_6$ and $\bx_7$ the outputs. Functions can take any number of inputs (e.g. $f_3$ and $f_5$ take two) and have any number of outputs (e.g. $f_4$ has two). There are a few noteworthy properties of this graph:

\begin{enumerate}
\item The graph is bipartite, in the sense that arrows always go from boxes to circles and from circles to boxes. 
\item Functions can have any number of inputs or outputs; variables and parameters can have an arbitrary number of outputs (a parameter with more of one output is \emph{shared} between different layers); variables have at most one input and parameters none. 
\item Variables with no incoming arrows and parameters are not computed by the network, but must be set prior to evaluation, i.e.\ they are \emph{inputs}. Any variable (or even parameter) may be used as output, although these are usually the variables with no outgoing arrows.
\item Since the graph is acyclic, the CNN can be evaluated by sorting the functions and computing them one after another (in the example, evaluating the functions in the order $f_1,f_2,f_3,f_4,f_5$ would work).
\end{enumerate}

\section{Computing derivatives with backpropagation}\label{s:back}

Learning a NN requires computing the derivative of the loss with respect to the network parameters. Derivatives are computed using an algorithm called \emph{backpropagation}, which is a memory-efficient implementation of the chain rule for derivatives. First, we discuss the derivatives of a single layer, and then of a whole network.

\subsection{Derivatives of tensor functions}

In a CNN, a layer is a function $\by = f(\bx)$ where both input $\bx \in \mathbb{R}^{H\times W \times C}$ and output $\by \in \mathbb{R}^{H'\times W' \times C'}$ are tensors. The derivative of the function $f$ contains the derivative of each output component $y_{i'j'k'}$ with respect to each input component $x_{ijk}$, for a total of $H'\times W'\times C'\times H\times W\times C$ elements naturally arranged in a 6D tensor. Instead of expressing derivatives as tensors, it is often useful  to switch to a matrix notation by \emph{stacking} the input and output tensors into vectors. This is done by the $\vv$ operator, which visits each element of a tensor in lexicographical order and produces a vector:
\[
  \vv \bx
  =
  \begin{bmatrix}
  x_{111} \\
  x_{211} \\
  \vdots
  \\
  x_{H11} \\
  x_{121} \\
  \vdots \\
  x_{HWC}  	
  \end{bmatrix}.
\]
By stacking both input and output, each layer $f$ can be seen reinterpreted as vector function $\vv f$, whose derivative is the conventional Jacobian matrix:
\[
\renewcommand*{\arraystretch}{1.5}
\frac{d \vv f}{d(\vv \bx)^\top}
=
\begin{bmatrix}
\frac{\partial y_{111}}{\partial x_{111}} & 
\frac{\partial y_{111}}{\partial x_{211}} &
\dots &
\frac{\partial y_{111}}{\partial x_{H11}} &
\frac{\partial y_{111}}{\partial x_{121}} &
\dots &
\frac{\partial y_{111}}{\partial x_{HWC}} \\
\frac{\partial y_{211}}{\partial x_{111}} & 
\frac{\partial y_{211}}{\partial x_{211}} &
\dots &
\frac{\partial y_{211}}{\partial x_{H11}} &
\frac{\partial y_{211}}{\partial x_{121}} &
\dots &
\frac{\partial y_{211}}{\partial x_{HWC}} \\
\vdots & \vdots & \dots & \vdots & \vdots & \dots & \vdots \\
\frac{\partial y_{H'11}}{\partial x_{111}} & 
\frac{\partial y_{H'11}}{\partial x_{211}} &
\dots &
\frac{\partial y_{H'11}}{\partial x_{H11}} &
\frac{\partial y_{H'11}}{\partial x_{121}} &
\dots &
\frac{\partial y_{H'11}}{\partial x_{HWC}} \\
\frac{\partial y_{121}}{\partial x_{111}} & 
\frac{\partial y_{121}}{\partial x_{211}} &
\dots &
\frac{\partial y_{121}}{\partial x_{H11}} &
\frac{\partial y_{121}}{\partial x_{121}} &
\dots &
\frac{\partial y_{121}}{\partial x_{HWC}} \\
\vdots & \vdots & \dots & \vdots & \vdots & \dots & \vdots \\
\frac{\partial y_{H'W'C'}}{\partial x_{111}} & 
\frac{\partial y_{H'W'C'}}{\partial x_{211}} &
\dots &
\frac{\partial y_{H'W'C'}}{\partial x_{H11}} &
\frac{\partial y_{H'W'C'}}{\partial x_{121}} &
\dots &
\frac{\partial y_{H'W'C'}}{\partial x_{HWC}}
\end{bmatrix}.
\]
This notation for the derivatives of tensor functions is taken from~\cite{kinghorn96integrals} and is used throughout this document.

While it is easy to express the derivatives of tensor functions as matrices, these matrices are in general extremely large. Even for moderate data sizes (e.g. $H=H'=W=W'=32$ and $C=C'=128$), there are $H'W'C'HWC \approx 17 \times 10^9$ elements in the Jacobian. Storing that requires 68 GB of space in single precision. The purpose of the backpropagation algorithm is to compute the derivatives required for learning without incurring this huge memory cost.

\subsection{Derivatives of function compositions}

In order to understand backpropagation, consider first a simple CNN terminating in a loss function $f_L = \ell_\by$:
\begin{center}
\begin{tikzpicture}[auto, node distance=2cm]
\node (x0)  [data] {$\bx_0$};
\node (f1) [block,right of=x0]{$f_1$};
\node (f2) [block,right of=f1,node distance=3cm]{$f_2$};
\node (dots) [right of=f2]{...};
\node (fL) [block,right of=dots]{$f_L$};
\node (w1) [data, below of=f1] {$\bw_1$};
\node (w2) [data, below of=f2] {$\bw_2$};
\node (wL) [data, below of=fL] {$\bw_L$};
\node (xL) [data, right of=fL] {$x_l\in\real$};
\draw [to] (x0.east) -- (f1.west) {};
\draw [to] (f1.east) -- node {$\bx_1$} (f2.west);
\draw [to] (f2.east) -- node {$\bx_2$} (dots.west) {};
\draw [to] (dots.east) -- node {$\bx_{L-1}$} (fL.west) {};
\draw [to] (fL.east) -- (xL.west) {};
\draw [to] (w1.north) -- (f1.south) {};
\draw [to] (w2.north) -- (f2.south) {};
\draw [to] (wL.north) -- (fL.south) {};
\end{tikzpicture}
\end{center}
The goal is to compute the gradient of the loss value $x_L$ (output) with respect to each network parameter $\bw_l$:
\[
\frac{df}{d(\vv \bw_l)^\top} = 
\frac{d}{d(\vv \bw_l)^\top}
\left[f_L(\cdot;\bw_L) \circ ... \circ 
f_2(\cdot;\bw_2) \circ f_1(\bx_0;\bw_1)\right].
\]
By applying the chain rule and by using the matrix notation introduced above, the derivative can be written as
\begin{equation}\label{e:chain-rule}
\frac{df}{d(\vv \bw_l)^\top} 
= 
\frac{d\vv f_L(\bx_{L-1};\bw_{L})}{d(\vv\bx_{L-1})^\top}
\times
\dots
\times
\frac{d\vv f_{l+1}(\bx_{l};\bw_{l+1})}{d(\vv\bx_{l})^\top}
\times
\frac{d\vv f_l(\bx_{l-1};\bw_{l})}{d(\vv\bw_l^\top)}
\end{equation}
where the derivatives are computed at the working point determined by the input $\bx_0$ and the current value of the parameters. 

Note that, since the network output $x_l$ is a \emph{scalar} quantity, the target derivative $df/d(\vv \bw_l)^\top$ has the same number of elements of the parameter vector $\bw_l$, which is moderate. However, the intermediate Jacobian factors have, as seen above, an unmanageable size. In order to avoid computing these factor explicitly, we can proceed as follows.

Start by multiplying the output of the last layer by a tensor $p_L=1$ (note that this tensor is a scalar just like the variable $x_L$):
\begin{align*}
p_L \times \frac{df}{d(\vv \bw_l)^\top} 
&= 
\underbrace{p_L \times \frac{d\vv f_L(\bx_{L-1};\bw_{L})}{d(\vv\bx_{L-1})^\top}}_{(\vv \bp_{L-1})^\top}
\times
\dots
\times
\frac{d\vv f_{l+1}(\bx_{l};\bw_{l+1})}{d(\vv\bx_{l})^\top}
\times
\frac{d\vv f_l(\bx_{l-1};\bw_{l})}{d(\vv\bw_l^\top)}
\\
&=
(\vv \bp_{L-1})^\top
\times
\dots
\times
\frac{d\vv f_{l+1}(\bx_{l};\bw_{l+1})}{d(\vv\bx_{l})^\top}
\times
\frac{d\vv f_l(\bx_{l-1};\bw_{l})}{d(\vv\bw_l^\top)}
\end{align*}
In the second line the last two factors to the left have been multiplied obtaining a new tensor $\bp_{L-1}$ that has the same size as the variable $\bx_{L-1}$. The factor $\bp_{L-1}$ can therefore be explicitly stored. The construction is then repeated by multiplying pairs of factors from left to right, obtaining a sequence of tensors $\bp_{L-2},\dots,\bp_{l}$ until the desired derivative is obtained. Note that, in doing so, no large tensor is ever stored in memory. This process is known as \emph{backpropagation}.

In general, tensor $\bp_{l}$ is obtained from $\bp_{l+1}$ as the product:
\[
(\vv \bp_{l})^\top = (\vv \bp_{l+1})^\top \times \frac{d\vv f_{l+1}(\bx_{l};\bw_{l+1})}{d(\vv\bx_{l})^\top}.
\]
The key to implement backpropagation is to be able to compute these products without explicitly computing and storing in memory the second factor, which is a large Jacobian matrix. Since computing the derivative is a linear operation, this product can be interpreted as the \emph{derivative of the layer projected along direction $\bp_{l+1}$}: 
\begin{equation}\label{e:projected}
\bp_{l} = 
\frac{d \langle \bp_{l+1}, f(\bx_l;\bw_l) \rangle}
{d \bx_{l}}.
\end{equation}
Here $\langle \cdot,\cdot \rangle$ denotes the inner product between tensors, which results in a scalar quantity. Hence the derivative \eqref{e:projected} needs not to use the $\vv$ notation, and yields a tensor $\bp_l$ that has the same size as $\bx_l$ as expected.

In order to implement backpropagation, a CNN toolbox provides implementations of each layer $f$ that provide:
\begin{itemize}
\item A \textbf{forward mode}, computing the output $\by = f(\bx;\bw)$ of the layer given its input $\bx$ and parameters $\bw$.
\item A \textbf{backward mode}, computing the projected derivatives
\[
\frac{d \langle \bp, f(\bx;\bw) \rangle}
{d \bx}
\quad\text{and}\quad
\frac{d \langle \bp, f(\bx;\bw) \rangle}
{d \bw},
\]
given, in addition to the input $\bx$ and parameters $\bw$, a tensor $\bp$ that the same size as $\by$.
\end{itemize}
This is best illustrated with an example. Consider a layer $f$ such as the convolution operator implemented by the \matconvnet\ !vl_nnconv! command. In the ``forward'' mode, one calls the function as !y = vl_nnconv(x,w,[])! to apply the filters !w! to the input !x! and obtain the output !y!. In the ``backward mode'', one calls ![dx, dw] = vl_nnconv(x,w,[],p)!.  As explained above, !dx!, !dw!, and !p! have the same size as !x!, !w!, and !y!, respectively. The computation of large Jacobian is encapsulated in the function call and never carried out explicitly. 

\subsection{Backpropagation networks}\label{s:bpnets}

In this section, we provide a schematic interpretation of backpropagation and show how it can be implemented by ``reversing'' the NN computational graph.

The projected derivative of eq.~\eqref{e:projected} can be seen as the derivative of the following mini-network:
\begin{center}
\begin{tikzpicture}[auto, node distance=2cm]
\node (x) [data] {$\bx$};
\node (f) [block,right of=x ] {$f$};
\node (dot)[block,right of=f ] {$\langle \cdot, \cdot \rangle$};
\node (z) [data, right of=dot] {$z \in \mathbb{R}$};
\node (w) [data, below of=f ] {$\bw$};
\node (p) [data, below of=dot] {$\bp$};
\draw [to] (x.east) -- (f.west) {};
\draw [to] (f.east) -- node {$\by$}  (dot.west) {};
\draw [to] (w.north) -- (f.south) {};
\draw [to] (dot.east) -- (z.west) {};
\draw [to] (p.north) -- (dot.south) {};
\end{tikzpicture}
\end{center}
In the context of back-propagation, it can be useful to think of the projection $\bp$ as the ``linearization'' of the rest of the network from variable $\by$ down to the loss. The projected derivative can also be though of as a new layer $(d\bx, d\bw) = df(\bx,\bw,\bp)$ that, by computing the derivative of the mini-network, operates in the reverse direction:
\begin{center}
\begin{tikzpicture}[auto, node distance=2cm]
\node (df) [block,right of=x] {$df$};
\node (dx) [data,left of=df] {$d\bx$};
\node (dw) [data,below of=df] {$d\bw$};
\node (w) [data,above of=df,xshift=0.6em] {$\bw$};
\node (x) [data,above of=df,xshift=-0.6em] {$\bx$};
\node (p) [data,right of=df] {$\bp$};
\draw [to] (df.west) -- (dx.east)  {};
\draw [to] (df.south) -- (dw.north)  {};
\draw [to] (p.west) -- (f.east) {};
\draw [to] (w.south) -- ([xshift=0.6em]df.north) {};
\draw [to] (x.south) -- ([xshift=-0.6em]df.north) {};
\end{tikzpicture}
\end{center}
By construction (see eq.~\eqref{e:projected}), the function $df$ is \emph{linear} in the argument $\bp$.

Using this notation, the forward and backward passes through the original network can be rewritten as evaluating an extended network which contains a BP-reverse of the original one (in blue in the diagram):
\begin{center}
\begin{tikzpicture}[auto, node distance=2cm]
\node (x0) [data] {$\bx_0$};
\node (f1) [block,right of=x0] {$f_1$};
\node (x1) [data,right of=f1] {$\bx_{1}$};
\node (w1) [data,below of=f1] {$\bw_1$};
\node (f2) [block,right of=x1] {$f_2$};
\node (x2) [data,right of=f2] {$\bx_{2}$};
\node (w2) [data,below of=f2] {$\bw_2$};
\node (f3) [right of=x2] {$\dots$};
\node (xLm) [right of=f3] {$\bx_{L-1}$};
\node (fL) [block,right of=xLm] {$f_L$};
\node (xL) [data,right of=fL] {$\bx_{L}$};
\node (wL) [data,below of=fL] {$\bw_L$};
\draw [to] (x0.east) -- (f1.west) {};
\draw [to] (w1.north) -- (f1.south) {};
\draw [to] (f1.east) -- (x1.west) {};
\draw [to] (x1.east) -- (f2.west) {};
\draw [to] (w2.north) -- (f2.south) {};
\draw [to] (f2.east) -- (x2.west) {};
\draw [to] (x2.east) -- (f3.west) {};
\draw [to] (f3.east) -- (xLm.west) {};
\draw [to] (xLm.east) -- (fL.west) {};
\draw [to] (wL.north) -- (fL.south) {};
\draw [to] (fL.east) -- (xL.west) {};
\node (dfL) [block,below of=wL,bp] {$df_L$};
\node (dxL) [data,right of=dfL,bpe] {$d\bp_L$};
\node (dwL) [data,below of=dfL,bpe] {$d\bw_L$};
\node (dxLm) [data,left of=dfL,bpe] {$d\bx_{L-1}$};
\node (df3) [left of=dxLm,bpe] {$\dots$};
\node (df2) [block,below of=w2,bp] {$df_2$};
\node (dx2) [data,right of=df2,bpe] {$d\bx_{2}$};
\node (dw2) [data,below of=df2,bpe] {$d\bw_2$};
\node (df1) [block,below of=w1,bp] {$df_1$};
\node (dx1) [data,right of=df1,bpe] {$d\bx_{1}$};
\node (dw1) [data,below of=df1,bpe] {$d\bw_1$};
\node (dx0) [data,left of=df1,bpe] {$d\bx_{0}$};
\draw [to,bp] (wL.south) -- (dfL.north) {};
\draw [to,bp] (dfL.south) -- (dwL.north) {};
\draw [to,bp] (dxL.west) -- (dfL.east) {};
\draw [to,bp] (dfL.west) -- (dxLm.east) {};
\draw [to,bp] (dxLm.west) -- (df3.east) {};
\draw [to,bp] (df3.west) -- (dx2.east) {};
\draw [to,bp] (w2.south) -- (df2.north) {};
\draw [to,bp] (df2.south) -- (dw2.north) {};
\draw [to,bp] (dx2.west) -- (df2.east) {};
\draw [to,bp] (df2.west) -- (dx1.east) {};
\draw [to,bp] (w1.south) -- (df1.north) {};
\draw [to,bp] (df1.south) -- (dw1.north) {};
\draw [to,bp] (dx1.west) -- (df1.east) {};
\draw [to,bp] (df1.west) -- (dx0.east) {};
\draw [to,bp] (x0) -- (df1) {} ;
\draw [to,bp] (x1) -- (df2) {} ;
\draw [to,bp] (xLm) -- (dfL) {} ;
\end{tikzpicture}
\end{center}

\subsection{Backpropagation in DAGs}\label{s:dag}

Assume that the DAG has a single output variable $\bx_L$ and assume, without loss of generality, that all variables are sorted in order of computation $(\bx_0,\bx_1,\dots,\bx_{L-1},\bx_L)$ according to the DAG structure. Furthermore, in order to simplify the notation, assume that this list contains both data and parameter variables, as the distinction is moot for the discussion in this section.

We can cut the DAG at any point in the sequence by fixing $\bx_0, \dots, \bx_{l-1}$ to some arbitrary value and dropping all the DAG layers that feed into them, effectively transforming the first $l$ variables into inputs. Then, the rest of the DAG defines a function $h_l$ that maps these input variables to the output $\bx_L$:
\[
 \bx_L = h_l(\bx_0,\bx_1,\dots,\bx_{l-1}).
\]
Next, we show that backpropagation in a DAG iteratively computes the projected derivatives of all functions $h_1,\dots,h_L$ with respect to all their parameters.

Backpropagation starts by initializing variables $(d\bx_{0},\dots,d\bx_{l-1})$ to null tensors of the same size as $(\bx_0,\dots,\bx_{l-1})$. Next, it computes the projected derivatives of
\[
 \bx_L = h_L(\bx_0,\bx_1,\dots,\bx_{L-1}) =
 f_{\pi_L}(\bx_0,\bx_1,\dots,\bx_{L-1}).
\]
Here $\pi_l$ denotes the index of the layer $f_{\pi_l}$ that computes the value of the variable $\bx_l$. There is at most one such layer, or none if $\bx_l$ is an input or parameter of the original NN. In the first case, the layer may depend on any of the variables prior to $\bx_l$ in the sequence, so that general one has:
\[
 \bx_{l} = f_{\pi_l}(\bx_0,\dots,\bx_{l-1}).
\]
	At the beginning of backpropagation, since there are no intermediate variables between $\bx_{L-1}$ and $\bx_L$, the function $h_L$ is the same as the last layer $f_{\pi_L}$. Thus the projected derivatives of $h_L$ are the same as the projected derivatives of $f_{\pi_L}$, resulting in the equation
\[
\forall t=0,\dots,L-1:\qquad
d\bx_{t} \leftarrow d\bx_{t}
+ \frac{d\langle \bp_L, f_{\pi_L}(\bx_0,\dots,\bx_{t-1})\rangle}{d\bx_t}.
\]
Here, for uniformity with the other iterations, we use the fact that $d\bx_l$ are initialized to zero an\emph{accumulate} the values instead of storing them. In practice, the update operation needs to be carried out only for the variables $\bx_l$ that are actual inputs to $f_{\pi_L}$, which is often a tiny fraction of all the variables in the DAG.

After the update, each $d\bx_t$ contains the projected derivative of function $h_L$ with respect to the corresponding variable:
\[
\forall t=0,\dots,L-1:\qquad
d\bx_t = \frac{d\langle \bp_L, h_L(\bx_0,\dots,\bx_{l-1})\rangle}{d\bx_t}.
\]
Given this information, the next iteration of backpropagation updates the variables to contain the projected derivatives of $h_{L-1}$ instead. In general, given the derivatives of $h_{l+1}$, backpropagation computes the derivatives of $h_{l}$ by using the relation
\[
 \bx_L
 = 
 h_{l}(\bx_0,\bx_1,\dots,\bx_{l-1})
 =
 h_{l+1}(\bx_0,\bx_1,\dots,\bx_{l-1},f_{\pi_L}(\bx_0,\dots,\bx_{l-1}))
\]
Applying the chain rule to this expression, for all $0\leq t \leq l-1$:
\[
\frac{d\langle \bp, h_l \rangle}{d(\vv \bx_t)^\top}
=
\frac{d\langle \bp, h_{l+1}\rangle}{d(\vv \bx_t)^\top}
+
\underbrace{\frac{d\langle \bp_L, h_{l+1}\rangle}{d(\vv \bx_l)^\top}}_{\vv d\bx_l}
\frac{d \vv f_{\pi_l}}{d(\vv \bx_t)^\top}.
\]
This yields the update equation
\begin{equation}\label{e:bp-update}	
\forall t=0,\dots,l-1:\qquad
d\bx_t \leftarrow d\bx_t + \frac{d\langle \bp_l, f_{\pi_l}(\bx_0,\dots,\bx_{l-1})\rangle}{d\bx_t},
\quad
\text{where\ }
\bp_l = d\bx_l.
\end{equation}
Once more, the update needs to be explicitly carried out only for the variables $\bx_t$ that are actual inputs of $f_{\pi_l}$. In particular, if $\bx_l$ is a data input or a parameter of the original neural network, then $\bx_l$ does not depend on any other variables or parameters and $f_{\pi_l}$ is a nullary function (i.e.\ a function with no arguments). In this case, the update does not do anything. 
After iteration $L-l+1$ completes, backpropagation remains with:
\begin{align*}
\forall t=0,\dots,l-1:&\qquad
d\bx_t
=
\frac{d\langle \bp_L, h_l(\bx_0,\dots,\bx_{l-1})\rangle}{d\bx_t}.
\end{align*}
Note that the derivatives for variables $\bx_t, l \leq t \leq L-1$ are not updated since $h_l$ does not depend on any of those. Thus, after all $L$ iterations are complete, backpropagation terminates with
\[
\forall l=1,\dots,L:\qquad
d\bx_{l-1}
=
\frac{d\langle \bp_L, h_{l}(\bx_0,\dots,\bx_{l-1})\rangle}{d\bx_{l-1}}.
\]
As seen above, functions $h_{l}$ are obtained from the original network $f$ by transforming variables $\bx_0,\dots,\bx_{l-1}$ into to inputs. If $\bx_{l-1}$ was already an input (data or parameter) of $f$, then the derivative $d\bx_{l-1}$ is applicable to $f$ as well.

Backpropagation can be summarized as follows:
\begin{center}
\fbox{\begin{minipage}{0.95\textwidth}
Given: a DAG neural network $f$ with a single output $\bx_L$, the values of all input variables (including the parameters), and the value of the projection $\bp_L$ (usually $\bx_L$ is a scalar and $\bp_L = p_L = 1$):
\begin{enumerate}
    \item Sort all variables by computation order $(\bx_0,\bx_1,\dots,\bx_L)$ according to the DAG.
    \item Perform a forward pass through the network to compute all the intermediate variable values.
    \item Initialize $(d\bx_0, \dots, d\bx_{L-1})$ to null tensors with the same size as the corresponding variables.
    \item For $l=L,L-1,\dots,2,1$:
  \begin{enumerate}
  \item Find the index $\pi_l$ of the layer $\bx_{l} = f_{\pi_l}(\bx_0,\dots,\bx_{l-1})$ that evaluates variable $\bx_l$. If there is no such layer (because $\bx_{l}$ is an input or parameter of the network), go to the next iteration.
  \item Update the variables using the formula:
   \[
   \forall t=0,\dots,l-1:\qquad
d\bx_t \leftarrow d\bx_t + \frac{d\langle d\bx_l, f_{\pi_l}(\bx_0,\dots,\bx_{l-1})\rangle}{d\bx_t}.
   \]
   To do so efficiently, use the ``backward mode'' of the layer $f_{\pi_l}$ to compute its derivative projected onto $d\bx_l$ as needed.
  \end{enumerate}
  \end{enumerate}
\end{minipage}}
\end{center}


\begin{figure}[t]
\begin{center}
\begin{tikzpicture}[auto, node distance=0.3cm]
 \matrix (m) [matrix of math nodes, 
    column sep=1.2cm,
    row sep=0.3cm]
{
& \node (f1) [block]{f_1}; 
& \node (x1) [datac]{\bx_1};
\\
\node (x0) [datac]{\bx_0};
&
&
& \node (f3) [block]{f_3};
& \node (x3) [datac]{\bx_3};
\\
& \node (f2) [block]{f_2}; 
& \node (x2) [datac]{\bx_2};
& &
& \node (f5) [block]{f_5}; 
& \node (x7) [datac]{\bx_7}; 
\\
& 
& \node(x5) [datac]{\bx_5};
\\
\node (x4) [datac]{\bx_4};
& \node (f4) [block]{f_4};
\\
& 
& \node(x6) [datac]{\bx_6};
\\
& \node (df1) [block,bp]{df_1}; 
& \node (dx1) [datac,bp]{d\bx_1};
\\
\node (dx0) [datac,bp]{d\bx_0};
&
&
& \node (df3) [block,bp]{df_3};
& \node (dx3) [datac,bp]{d\bx_3};
\\
& \node (df2) [block,bp]{df_2}; 
& \node (dx2) [datac,bp]{d\bx_2};
& &
& \node (df5) [block,bp]{df_5}; 
& \node (dx7) [datac,bp]{\bp_7}; 
\\
& 
& \node (dx5) [datac,bp]{d\bx_5};
\\
\node (dx4) [datac,bp]{d\bx_4};
& \node (df4) [block,bp]{df_4};
\\
& 
& \node(dx6) [datac,bp]{\bp_6};
\\
};
\draw[to] (x0) -- (f1);
\draw[to] (f1) -- (x1);
\draw[to] (x1) -- (f3);
\draw[to] (x0) -- (f2);
\draw[to] (f2) -- (x2);
\draw[to] (x2) -- (f3);
\draw[to] (f3) -- (x3);
\draw[to] (x3) -- (f5);
\draw[to] (f5) -- (x7);
\draw[to] (x4) -- (f4);
\draw[to] (f4) -- (x5);
\draw[to] (f4) -- (x6);
\draw[to] (x5) -- (f5);
\node(w1) [par,below=of f1]{$\bw_1$}; \draw[to] (w1) -- (f1);
\node(w2) [par,below=of f2]{$\bw_2$}; \draw[to] (w2) -- (f2);
\node(w4) [par,below=of f4]{$\bw_4$}; \draw[to] (w4) -- (f4);
\draw[to] (w4) to [bend right] (f3);
\node(w5) [par,below=of f5]{$\bw_5$}; \draw[to] (w5) -- (f5);
\node (dx0s) [right of=dx0,xshift=20pt,draw,rectangle,bp]{$\Sigma$};
\draw[from,bp] (dx0) -- (dx0s);
\draw[from,bp] (dx0s) -- (df1);
\draw[from,bp] (df1) -- (dx1);
\draw[from,bp] (dx1) -- (df3);
\draw[from,bp] (dx0s) -- (df2);
\draw[from,bp] (df2) -- (dx2);
\draw[from,bp] (dx2) -- (df3);
\draw[from,bp] (df3) -- (dx3);
\draw[from,bp] (dx3) -- (df5);
\draw[from,bp] (df5) -- (dx7);
\draw[from,bp] (dx4) -- (df4);
\draw[from,bp] (df4) -- (dx5);
\draw[from,bp] (df4) -- (dx6);
\draw[from,bp] (dx5) -- (df5);
\node(dw1) [par,below=of df1,bp]{$d\bw_1$}; \draw[from,bp] (dw1) -- (df1);
\node(dw2) [par,below=of df2,bp]{$d\bw_2$}; \draw[from,bp] (dw2) -- (df2);
\node(dw4s) [below of=df4,draw,rectangle,bp,yshift=-25pt]{$\Sigma$}; \draw[from,bp] (dw4s) -- (df4);
\node(dw4) [par,below=of dw4s,bp]{$d\bw_4$}; \draw[from,bp] (dw4) -- (dw4s);
\draw[from,bp] (dw4s) to [bend right,bp] (df3);
\node(dw5) [par,below=of df5,bp]{$d\bw_5$}; \draw[from,bp] (dw5) -- (df5);
\draw[to,bpl] (x0) -| ([xshift=-0.3cm]x0.west) |- (df1);
\draw[to,bpl] (x0) -| ([xshift=-0.6cm]x0.west) |- (df2);
\draw[to,bpl] (x1) -| ([xshift=4cm]x1.west) |- ([yshift=10pt]df3.east);
\draw[to,bpl] (x2) -| (df3);
\draw[to,bpl] (x3) -| ([xshift=+5cm]x3.east) |- ([yshift=15pt]df5.east);
\draw[to,bpl] (x4) to [bend right=75] ([yshift=15pt]df4.west);
\draw[to,bpl] (x5) to [bend left] (df5);
\end{tikzpicture}
\end{center}
\vspace{-1em}
\caption{\textbf{Backpropagation network for a DAG.}}\label{f:dagbp}
\end{figure}
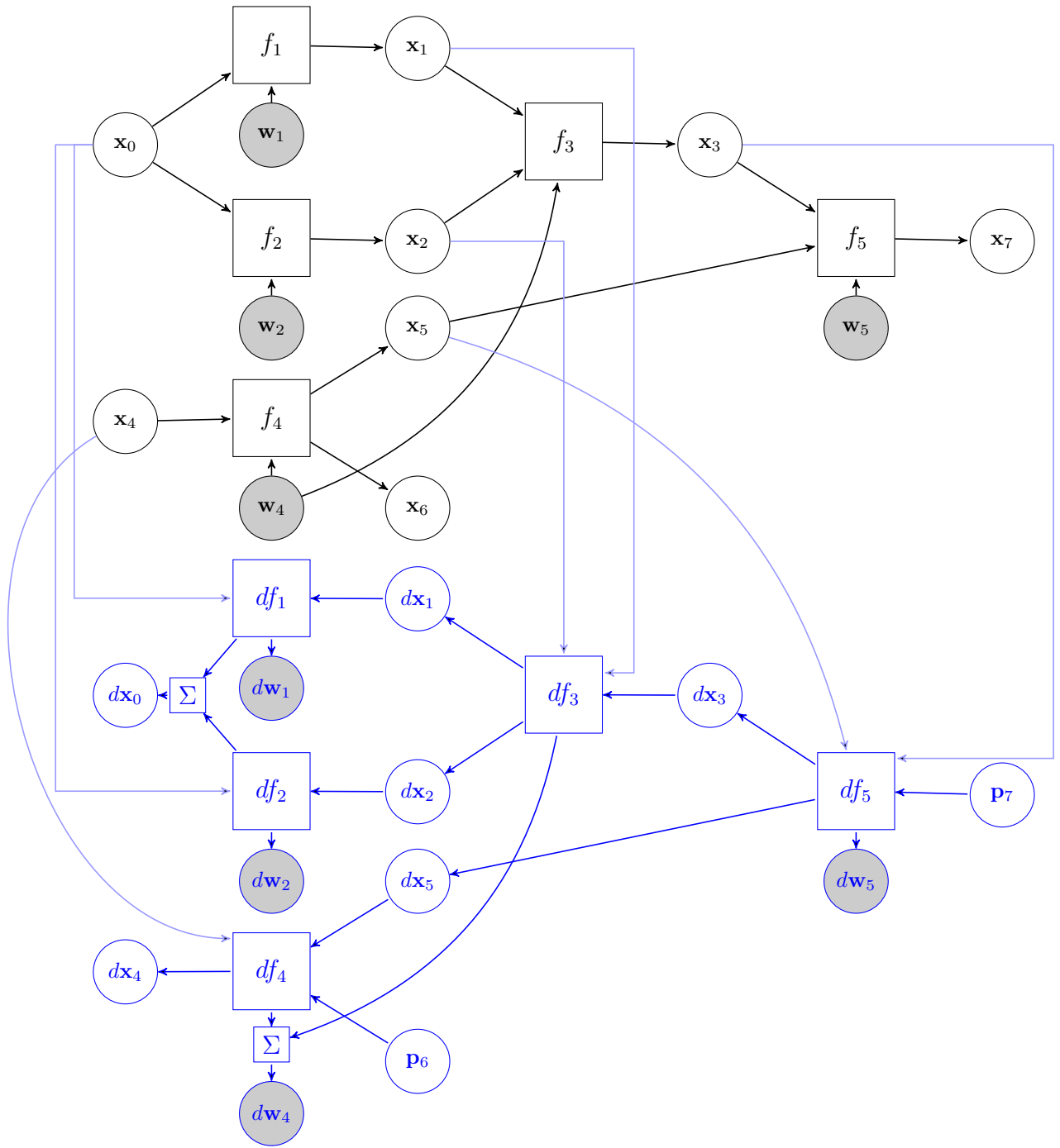

\subsection{DAG backpropagation networks}\label{s:bpnets-dag}

Just like for sequences, backpropagation in DAGs can be implemented as a corresponding BP-reversed DAG. To construct the reversed DAG:
\begin{enumerate}
\item For each layer $f_l$, and variable/parameter $\bx_t$ and $\bw_l$, create a corresponding layer $df_l$ and variable/parameter $d\bx_t$ and $d\bw_l$.
\item If a variable $\bx_t$ (or parameter $\bw_l$) is an input of $f_l$, then it is an input of $df_l$ as well.
\item If a variable $\bx_t$ (or parameter $\bw_l$) is an input of $f_l$, then the variable $d\bx_t$ (or the parameter $d\bw_l$) is an output $df_l$.
\item In the previous step, if a variable $\bx_t$ (or parameter $\bw_l$) is input to two or more layers in $f$, then $d\bx_t$ would be the output of two or more layers in the reversed network, which creates a conflict. Resolve these conflicts by inserting a summation layer that adds these contributions (this corresponds to the summation in the BP update equation \eqref{e:bp-update}).
\end{enumerate}
The BP network corresponding to the DAG of Fig.~\ref{f:dag} is given in Fig.~\ref{f:dagbp}.

\chapter{Wrappers and pre-trained models}\label{s:wrappers}

It is easy enough to combine the computational blocks of \cref{s:blocks} ``manually''. However, it is usually much more convenient to use them through a \emph{wrapper} that can implement CNN architectures given a model specification. The available wrappers are briefly summarised in \cref{s:wrappers-overview}.

\matconvnet also comes with many pre-trained models for image classification (most of which are trained on the ImageNet ILSVRC challenge), image segmentation, text spotting, and face recognition. These are very simple to use, as illustrated in \cref{s:pretrained}.

\section{Wrappers}\label{s:wrappers-overview}

\matconvnet provides two wrappers: SimpleNN for basic chains of blocks (\cref{s:simplenn}) and DagNN for blocks organized in more complex direct acyclic graphs (\cref{s:dagnn}).

\subsection{SimpleNN}\label{s:simplenn}

The SimpleNN wrapper is suitable for networks consisting of linear chains of computational blocks.  It is largely implemented by the \verb!vl_simplenn! function (evaluation of the CNN and of its derivatives), with a few other support functions such as \verb!vl_simplenn_move! (moving the CNN between CPU and GPU) and \verb!vl_simplenn_display! (obtain and/or print information about the CNN).

\verb!vl_simplenn! takes as input a structure \verb!net! representing the CNN as well as input \verb!x! and potentially output derivatives \verb!dzdy!, depending on the mode of operation. Please refer to the inline help of the \verb!vl_simplenn! function for details on the input and output formats. In fact, the implementation of \verb!vl_simplenn! is a good example of how the basic neural net building blocks can be used together and can serve as a basis for more complex implementations.

\subsection{DagNN}\label{s:dagnn}

The DagNN wrapper is more complex than SimpleNN as it has to support arbitrary graph topologies. Its design is object oriented, with one class implementing each layer type. While this adds complexity, and makes the wrapper slightly slower for tiny CNN architectures (e.g. MNIST), it is in practice much more flexible and easier to extend.

DagNN is implemented by the \verb!dagnn.DagNN! class (under the \verb!dagnn! namespace).

\section{Pre-trained models}\label{s:pretrained}

\verb!vl_simplenn! is easy to use with pre-trained models (see the homepage to download some). For example, the following code downloads a model pre-trained on the ImageNet data and applies it to one of MATLAB stock images:
\begin{lstlisting}[language=Matlab]
% setup MatConvNet in MATLAB
run matlab/vl_setupnn

% download a pre-trained CNN from the web
urlwrite(...
  'http://www.vlfeat.org/matconvnet/models/imagenet-vgg-f.mat', ...
  'imagenet-vgg-f.mat') ;
net = load('imagenet-vgg-f.mat') ;

% obtain and preprocess an image
im = imread('peppers.png') ;
im_ = single(im) ; % note: 255 range
im_ = imresize(im_, net.meta.normalization.imageSize(1:2)) ;
im_ = im_ - net.meta.normalization.averageImage ;
\end{lstlisting}
Note that the image should be preprocessed before running the network. While preprocessing specifics depend on the model, the pre-trained model contains a \verb!net.meta.normalization! field that describes the type of preprocessing that is expected. Note in particular that this network takes images of a fixed size as input and requires removing the mean; also, image intensities are normalized in the range [0,255].

The next step is running the CNN. This will return a \verb!res! structure with the output of the network layers:
\begin{lstlisting}[language=Matlab]
% run the CNN
res = vl_simplenn(net, im_) ;
\end{lstlisting}

The output of the last layer can be used to classify the image. The class names are contained in the \verb!net! structure for convenience:
\begin{lstlisting}[language=Matlab]
% show the classification result
scores = squeeze(gather(res(end).x)) ;
[bestScore, best] = max(scores) ;
figure(1) ; clf ; imagesc(im) ;
title(sprintf('%s (%d), score %.3f',...
net.meta.classes.description{best}, best, bestScore)) ;
\end{lstlisting}

Note that several extensions are possible. First, images can be cropped rather than rescaled. Second, multiple crops can be fed to the network and results averaged, usually for improved results. Third, the output of the network can be used as generic features for image encoding.

\section{Learning models}\label{s:wrappers-learning}

As \matconvnet can compute derivatives of the CNN using backpropagation, it is simple to implement learning algorithms with it. A basic implementation of stochastic gradient descent is therefore straightforward. Example code is provided in \verb!examples/cnn_train!. This code is flexible enough to allow training on NMINST, CIFAR, ImageNet, and probably many other datasets. Corresponding examples are provided in the \verb!examples/! directory.

\section{Running large scale experiments}

For large scale experiments, such as learning a network for ImageNet, a NVIDIA GPU (at least 6GB of memory) and adequate CPU and disk speeds are highly recommended. For example, to train on ImageNet, we suggest the following:
\begin{itemize}
\item Download the ImageNet data~\url{http://www.image-net.org/challenges/LSVRC}. Install it somewhere and link to it from \verb!data/imagenet12!
\item Consider preprocessing the data to convert all images to have a height of 256 pixels. This can be done with the supplied \verb!utils/preprocess-imagenet.sh! script. In this manner, training will not have to resize the images every time. Do not forget to point the training code to the pre-processed data.
\item Consider copying the dataset into a RAM disk (provided that you have enough memory) for faster access. Do not forget to point the training code to this copy.
\item Compile \matconvnet with GPU support. See the homepage for instructions.
\end{itemize}

Once your setup is ready, you should be able to run \verb!examples/cnn_imagenet! (edit the file and change any flag as needed to enable GPU support and image pre-fetching on multiple threads).

If all goes well, you should expect to be able to train with 200-300 images/sec.
\chapter{Computational blocks}\label{s:blocks}

This chapters describes the individual computational blocks supported by \matconvnet. The interface of a CNN computational block !<block>! is designed after the discussion in \cref{s:fundamentals}. The block is implemented as a MATLAB function !y = vl_nn<block>(x,w)! that takes as input MATLAB arrays !x! and !w! representing the input data and parameters and returns an array !y! as output. In general, !x! and !y! are 4D real arrays packing $N$ maps or images, as discussed above, whereas !w! may have an arbitrary shape.

The function implementing each block is capable of working in the backward direction as well, in order to compute derivatives. This is done by passing a third optional argument !dzdy! representing the derivative of the output of the network with respect to $\by$; in this case, the function returns the derivatives ![dzdx,dzdw] = vl_nn<block>(x,w,dzdy)! with respect to the input data and parameters. The arrays !dzdx!, !dzdy! and !dzdw! have the same dimensions of !x!, !y! and !w! respectively (see \cref{s:back}).

Different functions may use a slightly different syntax, as needed: many functions can take additional optional arguments, specified as property-value pairs; some do not have parameters  !w! (e.g. a rectified linear unit); others can take multiple inputs and parameters, in which case there may be more than one !x!, !w!, !dzdx!, !dzdy! or !dzdw!. See the rest of the chapter and MATLAB inline help for details on the syntax.\footnote{Other parts of the library will wrap these functions into objects with a perfectly uniform interface; however, the low-level functions aim at providing a straightforward and obvious interface even if this means differing slightly from block to block.}

The rest of the chapter describes the blocks implemented in \matconvnet, with a particular focus on their analytical definition. Refer instead to MATLAB inline help for further details on the syntax.

\section{Convolution}\label{s:convolution}

\begin{figure}[t]
	\centering
	\includegraphics[width=0.7\textwidth]{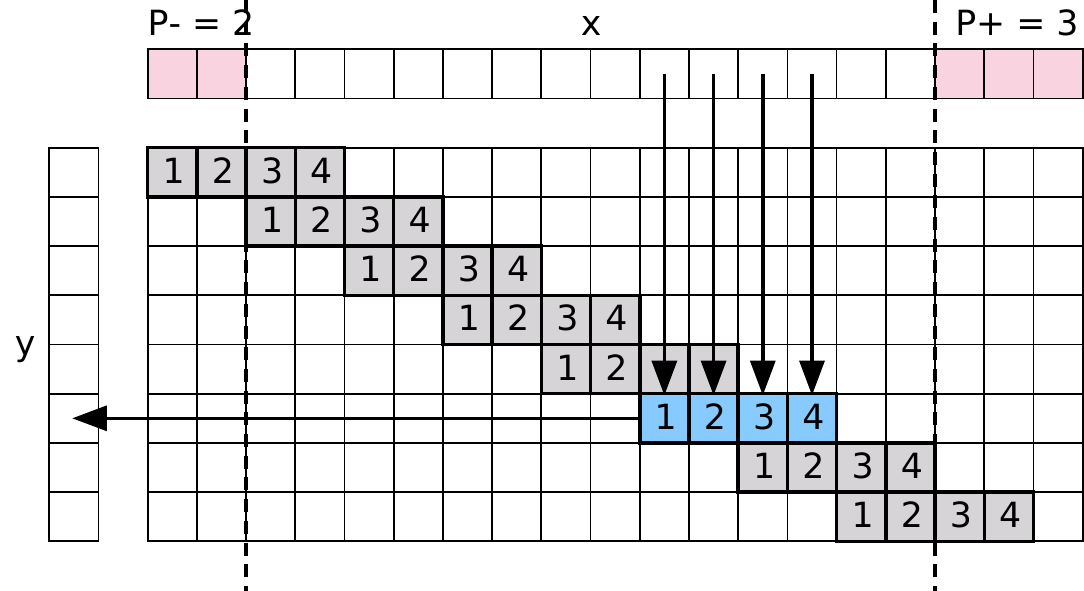}
	\caption{\textbf{Convolution.} The figure illustrates the process of filtering a 1D signal $\bx$ by a filter $f$ to obtain a signal $\by$. The filter has $H'=4$ elements and is applied with a stride of $S_h =2$ samples. The purple areas represented padding $P_-=2$ and $P_+=3$ which is zero-filled. Filters are applied in a sliding-window manner across the input signal. The samples of $\bx$ involved in the calculation of a sample of $\by$ are shown with arrow. Note that the rightmost sample of $\bx$  is never processed by any filter application due to the sampling step. While in this case the sample is in the padded region, this can happen also without padding.}\label{f:conv}
\end{figure}

The convolutional block is implemented by the function !vl_nnconv!. !y=vl_nnconv(x,f,b)! computes the convolution of the input map $\bx$ with a bank of $K$ multi-dimensional filters $\bff$ and biases $b$. Here
\[
 \bx\in\real^{H \times W \times D}, \quad
 \bff\in\real^{H' \times W' \times D \times D''}, \quad
 \by\in\real^{H'' \times W'' \times D''}.
\]
The process of convolving a signal is illustrated in \cref{f:conv} for a 1D slice. Formally, the output is given by
\[
y_{i''j''d''}
=
b_{d''}
+
\sum_{i'=1}^{H'}
\sum_{j'=1}^{W'}
\sum_{d'=1}^D
f_{i'j'd} \times x_{i''+i'-1,j''+j'-1,d',d''}.
\]
The call !vl_nnconv(x,f,[])! does not use the biases. Note that the function works with arbitrarily sized inputs and filters (as opposed to, for example, square images). See \cref{s:impl-convolution} for technical details.

\paragraph{Padding and stride.} !vl_nnconv! allows to specify  top-bottom-left-right paddings $(P_h^-,P_h^+,P_w^-,P_w^+)$ of the input array and subsampling strides $(S_h,S_w)$ of the output array:
\[
y_{i''j''d''}
=
b_{d''}
+
\sum_{i'=1}^{H'}
\sum_{j'=1}^{W'}
\sum_{d'=1}^D
f_{i'j'd} \times x_{S_h (i''-1)+i'-P_h^-, S_w(j''-1)+j' - P_w^-,d',d''}.
\]
In this expression, the array $\bx$ is implicitly extended with zeros as needed.

\paragraph{Output size.} !vl_nnconv! computes only the ``valid'' part of the convolution; i.e. it requires each application of a filter to be fully contained in the input support.  The size of the output is computed in \cref{s:receptive-simple-filters} and is given by:
\[
  H'' = 1 + \left\lfloor \frac{H - H' + P_h^- + P_h^+}{S_h} \right\rfloor.
\]
Note that the padded input must be at least as large as the filters: $H +P_h^- + P_h^+ \geq H'$, otherwise an error is thrown.

\paragraph{Receptive field size and geometric transformations.} Very often it is useful to geometrically relate the indexes of the various array to the input data (usually images) in terms of coordinate transformations and size of the receptive field (i.e. of the image region that affects an output). This is derived in \cref{s:receptive-simple-filters}.

\paragraph{Fully connected layers.} In other libraries, \emph{fully connected blocks or layers} are linear functions where each output dimension depends on all the input dimensions. \matconvnet does not distinguish between fully connected layers and convolutional blocks. Instead, the former is a special case of the latter obtained when the output map $\by$ has dimensions $W''=H''=1$. Internally, !vl_nnconv! handles this case more efficiently when possible.

\paragraph{Filter groups.} For additional flexibility, !vl_nnconv! allows to group channels of the input array $\bx$ and apply different subsets of filters to each group. To use this feature, specify as input a bank  of $D''$ filters $\bff\in\real^{H'\times W'\times D'\times D''}$ such that $D'$ divides the number of input dimensions $D$. These are treated as $g=D/D'$ filter groups; the first group is applied to dimensions $d=1,\dots,D'$ of the input $\bx$; the second group to dimensions $d=D'+1,\dots,2D'$ and so on. Note that the output is still an array $\by\in\real^{H''\times W''\times D''}$.

An application of grouping is implementing the Krizhevsky and Hinton network~\cite{krizhevsky12imagenet} which uses two such streams. Another application is sum pooling; in the latter case, one can specify $D$ groups of $D'=1$ dimensional filters identical filters of value 1 (however, this is considerably slower than calling the dedicated pooling function as given in \cref{s:pooling}).

\section{Convolution transpose (deconvolution)}\label{s:convt}

\begin{figure}[t]
	\centering
	\includegraphics[width=0.7\textwidth]{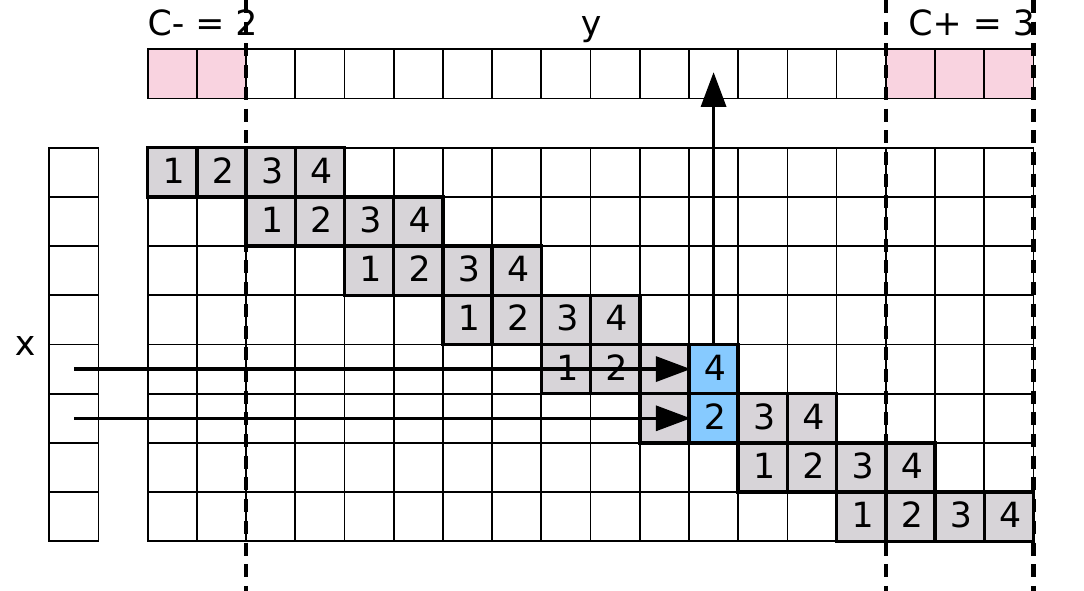}
	\caption{\textbf{Convolution transpose.} The figure illustrates the process of filtering a 1D signal $x$ by a filter $f$ to obtain a signal $y$. The filter is applied in a sliding-window, in a pattern that is the transpose of \cref{f:conv}. The filter has $H'=4$ samples in total, although each filter application uses two of them (blue squares) in a circulant manner. The purple areas represent crops with $C_-=2$ and $C_+=3$ which are discarded.  The samples of $x$ involved in the calculation of a sample of $y$ are shown with arrow. Note that, differently from \cref{f:conv}, there are no samples to the right of $\by$ which are involved in a convolution operation. This is because the width $H''$ of the output $\by$, which given $H'$ can be determined up to $U_h$ samples, is selected to be the smallest possible.}\label{f:convt}
\end{figure}

The \emph{convolution transpose} block (sometimes referred to as ``deconvolution'') is the transpose of the convolution block described in \cref{s:convolution}. In \matconvnet, convolution transpose is  implemented by the function !vl_nnconvt!.

In order to understand convolution transpose, let:
\[
 \bx\in\real^{H \times W \times D}, \quad
 \bff\in\real^{H' \times W' \times D \times D''}, \quad
 \by\in\real^{H'' \times W'' \times D''}, \quad
\]
be the input tensor, filters, and output tensors. Imagine operating in the reverse direction by using the filter bank $\bff$ to convolve the output $\by$ to obtain the input $\bx$, using the definitions given in~\cref{s:convolution} for the convolution operator; since convolution is linear, it can be expressed as a matrix $M$ such that  $\vv \bx = M \vv\by$; convolution transpose computes instead $\vv \by = M^\top \vv \bx$. This process is illustrated for a 1D slice in \cref{f:convt}.

There are two important applications of convolution transpose. The first one are the so called \emph{deconvolutional networks}~\cite{zeiler14visualizing} and other networks such as convolutional decoders that use the transpose of a convolution. The second one is implementing data interpolation. In fact, as the convolution block supports input padding and output downsampling, the convolution transpose block supports input upsampling and output cropping.

Convolution transpose can be expressed in closed form in the following rather unwieldy expression (derived in \cref{s:impl-convolution-transpose}):
\begin{multline}\label{e:convt}
y_{i''j''d''} =
 \sum_{d'=1}^{D}
 \sum_{i'=0}^{q(H',S_h)}
 \sum_{j'=0}^{q(W',S_w)}
f_{
1+ S_hi' + m(i''+ P_h^-, S_h),\ %
1+ S_wj' + m(j''+ P_w^-, S_w),\ %
d'',
d'
}
\times \\
x_{
1 - i' + q(i''+P_h^-,S_h),\ %
1 - j' + q(j''+P_w^-,S_w),\ %
d'
}
\end{multline}
where
\[
m(k,S) = (k - 1) \bmod S,
\qquad
q(k,n) = \left\lfloor \frac{k-1}{S} \right\rfloor,
\]
$(S_h,S_w)$ are the vertical and horizontal \emph{input upsampling factors},  $(P_h^-,P_h^+,P_h^-,P_h^+)$ the \emph{output crops}, and $\bx$ and $\bff$ are zero-padded as needed in the calculation. Note also that filter $k$ is stored as a slice $\bff_{:,:,k,:}$ of the 4D tensor $\bff$.

The height of the output array $\by$ is given by
\[
  H'' = S_h (H - 1) + H' -P^-_h - P^+_h.
\]
A similar formula holds true for the width. These formulas are derived in \cref{s:receptive-convolution-transpose} along with an expression for the receptive field of the operator.

We now illustrate the action of convolution transpose in an example (see also \cref{f:convt}).  Consider a 1D slice in the vertical direction, assume that the crop parameters are zero, and that $S_h>1$. Consider the output sample $y_{i''}$ where the index $i''$ is chosen such that $S_h$ divides $i''-1$; according to~\eqref{e:convt}, this sample is obtained as a weighted summation of $x_{i'' / S_h},x_{i''/S_h-1},...$ (note that the order is reversed). The weights are the filter elements $f_1$, $f_{S_h}$,$f_{2S_h},\dots$ subsampled with a step of $S_h$. Now consider computing the element $y_{i''+1}$; due to the rounding in the quotient operation $q(i'',S_h)$, this output sample is obtained as a weighted combination of the same elements of the input $x$ that were used to compute $y_{i''}$; however, the filter weights are now shifted by one place to the right: $f_2$, $f_{S_h+1}$,$f_{2S_h+1}$, $\dots$. The same is true for $i''+2, i'' + 3,\dots$ until we hit $i'' + S_h$. Here the cycle restarts after shifting $\bx$ to the right by one place. Effectively, convolution transpose works as an \emph{interpolating filter}.

\section{Spatial pooling}\label{s:pooling}

!vl_nnpool! implements max and sum pooling. The \emph{max pooling} operator computes the maximum response of each feature channel in a $H' \times W'$ patch
\[
y_{i''j''d} = \max_{1\leq i' \leq H', 1 \leq j' \leq W'} x_{i''+i-1',j''+j'-1,d}.
\]
resulting in an output of size $\by\in\real^{H''\times W'' \times D}$, similar to the convolution operator of \cref{s:convolution}. Sum-pooling computes the average of the values instead:
\[
y_{i''j''d} = \frac{1}{W'H'}
\sum_{1\leq i' \leq H', 1 \leq j' \leq W'} x_{i''+i'-1,j''+j'-1,d}.
\]
Detailed calculation of the derivatives is provided in \cref{s:impl-pooling}.

\paragraph{Padding and stride.} Similar to the convolution operator of \cref{s:convolution}, !vl_nnpool! supports padding the input; however, the effect is different from padding in the convolutional block as pooling regions straddling the image boundaries are cropped. For max pooling, this is equivalent to extending the input data with $-\infty$; for sum pooling, this is similar to padding with zeros, but the normalization factor at the boundaries is smaller to account for the smaller integration area.

\section{Activation functions}\label{s:activation}

\matconvnet supports the following activation functions:
\begin{itemize}
\item \emph{ReLU.} !vl_nnrelu! computes the \emph{Rectified Linear Unit} (ReLU):
\[
 y_{ijd} = \max\{0, x_{ijd}\}.
\]

\item \emph{Sigmoid.} !vl_nnsigmoid! computes the \emph{sigmoid}:
\[
 y_{ijd} = \sigma(x_{ijd}) = \frac{1}{1+e^{-x_{ijd}}}.
\]
\end{itemize}
See \cref{s:impl-activation} for implementation details.

\section{Spatial bilinear resampling}\label{s:spatial-sampler}

!vl_nnbilinearsampler! uses bilinear interpolation to spatially warp the image according to an input transformation grid. This operator works with an input image $\bx$, a grid $\bg$, and an output image $\by$ as follows:
\[
  \bx \in \mathbb{R}^{H \times W \times C},
  \qquad
  \bg \in [-1,1]^{2 \times H' \times W'},
  \qquad
  \by \in \mathbb{R}^{H' \times W' \times C}.
\]
The same transformation is applied to all the features channels in the input, as follows:
\begin{equation}\label{e:bilinear}
  y_{i''j''c}
  =
  \sum_{i=1}^H
  \sum_{j=1}^W
  x_{ijc}
  \max\{0, 1-|\alpha_v g_{1i''j''} + \beta_v - i|\}
  \max\{0, 1-|\alpha_u g_{2i''j''} + \beta_u - j|\},
\end{equation}
where, for each feature channel $c$, the output $y_{i''j''c}$ at the location $(i'',j'')$, is a weighted sum of the input values $x_{ijc}$ in the neighborhood of location $(g_{1i''j''},g_{2i''j''})$. The weights, as given in \eqref{e:bilinear}, correspond to performing bilinear interpolation. Furthermore, the grid coordinates are expressed not in pixels, but relative to a reference frame that extends from $-1$ to $1$ for all spatial dimensions of the input image; this is given by choosing the coefficients as:
\[
\alpha_v = \frac{H-1}{2},\quad
\beta_v = -\frac{H+1}{2},\quad
\alpha_u = \frac{W-1}{2},\quad
\beta_u = -\frac{W+1}{2}.
\]

See \cref{s:impl-sampler} for implementation details.

\section{Normalization}\label{s:normalization}

\subsection{Local response normalization (LRN)}\label{s:ccnormalization}

!vl_nnnormalize! implements the Local Response Normalization (LRN) operator. This operator is applied independently at each spatial location and to groups of feature channels as follows:
\[
 y_{ijk} = x_{ijk} \left( \kappa + \alpha \sum_{t\in G(k)} x_{ijt}^2 \right)^{-\beta},
\]
where, for each output channel $k$, $G(k) \subset \{1, 2, \dots, D\}$ is a corresponding subset of input channels. Note that input $\bx$ and output $\by$ have the same dimensions. Note also that the operator is applied uniformly at all spatial locations.

See \cref{s:impl-ccnormalization} for implementation details.

\subsection{Batch normalization}\label{s:bnorm}

!vl_nnbnorm! implements batch normalization~\cite{ioffe2015}. Batch normalization is somewhat different from other neural network blocks in that it performs computation across images/feature maps in a batch (whereas most blocks process different images/feature maps individually). !y = vl_nnbnorm(x, w, b)! normalizes each channel of the feature map $\mathbf{x}$ averaging over spatial locations and batch instances. Let $T$ be the batch size; then
\[
\mathbf{x}, \mathbf{y} \in \mathbb{R}^{H \times W \times K \times T},
\qquad\mathbf{w} \in \mathbb{R}^{K},
\qquad\mathbf{b} \in \mathbb{R}^{K}.
\]
Note that in this case the input and output arrays are explicitly treated as 4D tensors in order to work with a batch of feature maps. The tensors  $\mathbf{w}$ and $\mathbf{b}$ define component-wise multiplicative and additive constants. The output feature map is given by
\[
y_{ijkt} = w_k \frac{x_{ijkt} - \mu_{k}}{\sqrt{\sigma_k^2 + \epsilon}} + b_k,
\quad
\mu_{k} = \frac{1}{HWT}\sum_{i=1}^H \sum_{j=1}^W \sum_{t=1}^{T} x_{ijkt},
\quad
\sigma^2_{k} = \frac{1}{HWT}\sum_{i=1}^H \sum_{j=1}^W \sum_{t=1}^{T} (x_{ijkt} - \mu_{k})^2.
\]
See \cref{s:impl-bnorm} for implementation details.

\subsection{Spatial normalization}\label{s:spnorm}

!vl_nnspnorm! implements spatial normalization. The spatial normalization operator acts on different feature channels independently and rescales each input feature by the energy of the features in a local neighbourhood . First, the energy of the features in a neighbourhood $W'\times H'$ is evaluated
\[
n_{i''j''d}^2 = \frac{1}{W'H'}
\sum_{1\leq i' \leq H', 1 \leq j' \leq W'} x^2_{
i''+i'-1-\lfloor \frac{H'-1}{2}\rfloor,
j''+j'-1-\lfloor \frac{W'-1}{2}\rfloor,
d}.
\]
In practice, the factor $1/W'H'$ is adjusted at the boundaries to account for the fact that neighbors must be cropped. Then this is used to normalize the input:
\[
y_{i''j''d} = \frac{1}{(1 + \alpha n_{i''j''d}^2)^\beta} x_{i''j''d}.
\]
See \cref{s:impl-spnorm} for implementation details.

\subsection{Softmax}\label{s:softmax}

!vl_nnsoftmax! computes the softmax operator:
\[
 y_{ijk} = \frac{e^{x_{ijk}}}{\sum_{t=1}^D e^{x_{ijt}}}.
\]
Note that the operator is applied across feature channels and in a convolutional manner at all spatial locations. Softmax can be seen as the combination of an activation function (exponential) and a normalization operator. See \cref{s:impl-softmax} for implementation details.

\section{Categorical losses}\label{s:losses}

The purpose of a categorical loss function $\ell(\bx,\bc)$ is to compare a prediction $\bx$ to a ground truth class label $\bc$. As in the rest of \matconvnet, the loss is treated as a convolutional operator, in the sense that the loss is evaluated independently at each spatial location. However, the contribution of different samples are summed together (possibly after weighting) and the output of the loss is a scalar. \Cref{s:loss-classification} losses useful for multi-class classification and the \cref{s:loss-attributes} losses useful for binary attribute prediction. Further technical details are in \cref{s:impl-losses}. !vl_nnloss! implements the following all of these.

\subsection{Classification losses}\label{s:loss-classification}

Classification losses decompose additively as follows:
\begin{equation}\label{e:addloss}
\ell(\bx,\bc) = \sum_{ijn} w_{ij1n} \ell(\bx_{ij:n}, \bc_{ij:n}).
\end{equation}
Here $\bx \in \mathbb{R}^{H \times W \times C \times N}$ and $\bc \in \{1, \dots, C\}^{H \times W \times 1 \times N}$, such that the slice $\bx_{ij:n}$ represent a vector of $C$ class scores and and $c_{ij1n}$ is the ground truth class label. The !`instanceWeights`! option can be used to specify the tensor $\bw$ of weights, which are otherwise set to all ones; $\bw$ has the same dimension as $\bc$.

Unless otherwise noted, we drop the other indices and denote by $\bx$ and $c$  the slice $\bx_{ij:n}$ and the scalar $c_{ij1n}$. !vl_nnloss! automatically skips all samples such that $c=0$, which can be used as an ``ignore'' label.

\paragraph{Classification error.} The classification error is zero if class $c$ is assigned the largest score and zero otherwise:
\begin{equation}\label{e:loss-classerror}
\ell(\bx,c) = \mathbf{1}\left[c \not= \argmax_k x_c\right].
\end{equation}
Ties are broken randomly.

\paragraph{Top-$K$ classification error.} The top-$K$ classification error is zero if class $c$ is within the top $K$ ranked scores:
\begin{equation}\label{e:loss-classerror}
\ell(\bx,c) = \mathbf{1}\left[ |\{k : x_k \geq x_c \}| \leq K \right].
\end{equation}
The classification error is the same as the top-$1$ classification error.

\paragraph{Log loss or negative posterior log-probability.} In this case, $\bx$ is interpreted as a vector of posterior probabilities $p(k) = x_k, k=1,\dots, C$ over the $C$ classes. The loss is the negative log-probability of the ground truth class:
\begin{equation}\label{e:loss-log}
	\ell(\bx, c) = - \log x_c.
\end{equation}
Note that this makes the implicit assumption $\bx \geq 0, \sum_k x_k = 1$. Note also that, unless $x_c > 0$, the loss is undefined. For these reasons, $\bx$ is usually the output of a block such as softmax that can guarantee these conditions. However, the composition of the naive log loss and softmax is numerically unstable. Thus this is implemented as a special case below.

Generally, for such a loss to make sense, the score $x_c$ should be somehow in competition with the other scores $x_k, k\not = c$. If this is not the case, minimizing \eqref{e:loss-log} can trivially be achieved by maxing all $x_k$ large, whereas the intended effect is that $x_c$ should be large compared to the $x_k, k\not=c$. The softmax block makes the score compete through the normalization factor.

\paragraph{Softmax log-loss or multinomial logistic loss.} This loss combines the softmax block and the log-loss block into a single block:
\begin{equation}\label{e:loss-softmaxlog}
	\ell(\bx, c) = - \log \frac{e^{x_c}}{\sum_{k=1}^C e^{x_k}}
	= - x_c + \log \sum_{k=1}^C e^{x_k}.
\end{equation}
Combining the two blocks explicitly is required for numerical stability. Note that, by combining the log-loss with softmax, this loss automatically makes the score compete: $\ell(bx,c) \approx 0$ when $x_c \gg \sum_{k\not= c} x_k$.

This loss is implemented also in the \emph{deprecated} function !vl_softmaxloss!.

\paragraph{Multi-class hinge loss.} The multi-class logistic loss is given by
\begin{equation}\label{e:loss-multiclasshinge}
	\ell(\bx, c) = \max\{0, 1 - x_c \}.
\end{equation}
Note that $\ell(\bx,c) =0 \Leftrightarrow x_c \geq 1$. This, just as for the log-loss above, this loss does not automatically make the score competes. In order to do that, the loss is usually preceded by the block:
\[
 y_c = x_c - \max_{k \not= c} x_k.
\]
Hence $y_c$ represent the \emph{confidence margin} between class $c$ and the other classes $k \not= c$. Just like softmax log-loss combines softmax and loss, the next loss combines margin computation and hinge loss.

\paragraph{Structured multi-class hinge loss.} The structured multi-class logistic loss, also know as Crammer-Singer loss, combines the multi-class hinge loss with a block computing the score margin:
\begin{equation}\label{e:loss-structuredmulticlasshinge}
	\ell(\bx, c) = \max\left\{0, 1 - x_c + \max_{k \not= c} x_k\right\}.
\end{equation}

\subsection{Attribute losses}\label{s:loss-attributes}

Attribute losses are similar to classification losses, but in this case classes are not mutually exclusive; they are, instead, binary attributes. Attribute losses decompose additively as follows:
\begin{equation}\label{e:addlossattribute}
\ell(\bx,\bc) = \sum_{ijkn} w_{ijkn} \ell(\bx_{ijkn}, \bc_{ijkn}).
\end{equation}
Here $\bx\in \mathbb{R}^{H \times W \times C \times N}$ and $\bc \in \{-1,+1\}^{H \times W \times C \times N}$, such that the scalar $x_{ijkn}$ represent a confidence that attribute $k$ is on and $c_{ij1n}$ is the ground truth attribute label. The !`instanceWeights`! option can be used to specify the tensor $\bw$ of weights, which are otherwise set to all ones; $\bw$ has the same dimension as $\bc$.

 Unless otherwise noted, we drop the other indices and denote by $x$ and $c$  the scalars $x_{ijkn}$ and  $c_{ijkn}$. As before, samples with $c=0$ are skipped.

\paragraph{Binary error.} This loss is zero only if the sign of $x - \tau$ agrees with the ground truth label $c$:
\begin{equation}\label{e:loss-binary}
 \ell(x,c|\tau) = \mathbf{1}[\sign(x-\tau) \not= c].
\end{equation}
Here $\tau$ is a configurable threshold, often set to zero.

\paragraph{Binary log-loss.} This is the same as the multi-class log-loss but for binary attributes. Namely, this time $x_k \in [0,1]$ is interpreted as the probability that attribute $k$ is on:
\begin{align}\label{e:loss-binarylogloss}
\ell(x,c)
&=
\begin{cases}
- \log x, & c = +1, \\
- \log (1 - x), & c = -1, \\
\end{cases}
\\
&=
- \log \left[ c \left(x - \frac{1}{2}\right) + \frac{1}{2} \right].
\end{align}
Similarly to the multi-class log loss, the assumption $x \in [0,1]$ must be enforced by the block computing $x$.

\paragraph{Binary logistic loss.} This is the same as the multi-class logistic loss, but this time $x/2$ represents the confidence that the attribute is on and $-x/2$ that it is off. This is obtained by using the logistic function $\sigma(x)$
\begin{equation}\label{e:loss-binarylogistic}
 \ell(x,c)
 =
 - \log \sigma(cx)
 =
 -\log \frac{1}{1 + e^{-{cx}}}
 =
 -\log \frac{e^{\frac{cx}{2}}}{e^{\frac{cx}{2}} + e^{-\frac{cx}{2}}}.
\end{equation}

\paragraph{Binary hinge loss.} This is the same as the structured multi-class hinge loss but for binary attributes:
\begin{equation}\label{e:loss-hinge}
\ell(x,c)
=
\max\{0, 1 - cx\}.
\end{equation}
There is a relationship between the hinge loss and the structured multi-class hinge loss which is analogous to the relationship between binary logistic loss and multi-class logistic loss. Namely, the hinge loss can be rewritten as:
\[
\ell(x,c) = \max\left\{0, 1 - \frac{cx}{2} + \max_{k\not= c} \frac{kx}{2}\right\}
\]
Hence the hinge loss is the same as the structure multi-class hinge loss for $C=2$ classes, where $x/2$ is the score associated to class $c=1$ and $-x/2$ the score associated to class $c=-1$.

\section{Comparisons}\label{s:comparisons}

\subsection{$p$-distance}\label{s:pdistance}

The !vl_nnpdist! function computes the $p$-distance between the vectors in the input data $\bx$ and a target $\bar\bx$:
\[
  y_{ij} = \left(\sum_d |x_{ijd} - \bar x_{ijd}|^p\right)^\frac{1}{p}
\]
Note that this operator is applied convolutionally, i.e. at each spatial location $ij$ one extracts and compares vectors $x_{ij:}$. By specifying the option !'noRoot', true! it is possible to compute a variant omitting the root:
\[
  y_{ij} = \sum_d |x_{ijd} - \bar x_{ijd}|^p, \qquad p > 0.
\]
See \cref{s:impl-pdistance} for implementation details.

%
%
%
%
%
%

\chapter{Geometry}\label{s:geometry}

This chapter looks at the geometry of the CNN input-output mapping.

\section{Preliminaries}\label{s:preliminaries}

In this section we are interested in understanding how components in a CNN depend on components in the layers before it, and in particular on components of the input.  Since CNNs can incorporate blocks that perform complex operations, such as for example cropping their inputs based on data-dependent terms (e.g. Fast R-CNN), this information is generally available only at ``run time'' and cannot be uniquely determined given only the structure of the network. Furthermore, blocks can implement complex operations that are difficult to characterise in simple terms. Therefore, the analysis will be necessarily limited in scope.

We consider blocks such as convolutions for which one can deterministically establish dependency chains between network components. We also assume that all the inputs $\bx$ and outputs $\by$ are in the usual form of spatial maps, and therefore indexed as $x_{i,j,d,k}$ where $i,j$ are spatial coordinates.

Consider a layer $\by = f(\bx)$. We are interested in establishing which components of $\bx$ influence which components of $\by$. We also assume that this relation can be expressed in terms of a sliding rectangular window field, called \emph{receptive field}. This means that the output component  $y_{i'', j''}$ depends only on the input components $x_{i,j}$ where $(i,j) \in \Omega(i'', j'') $ (note that feature channels are implicitly coalesced in this discussion). The set $\Omega(i'',j'')$ is a rectangle defined as follows:
\begin{align}\label{e:receptive}
     i &\in \alpha_h (i'' -1) + \beta_h + \left[- \frac{\Delta_h-1}{2}, \frac{\Delta_h-1}{2}\right] \\
     j &\in \alpha_v (j'' -1) + \beta_v + \left[- \frac{\Delta_v-1}{2}, \frac{\Delta_v-1}{2}\right]
\end{align}
where $(\alpha_h,\alpha_v)$ is the \emph{stride}, $(\beta_h,\beta_v)$ the offset, and $(\Delta_h, \Delta_v)$ the \emph{receptive field size}.

\section{Simple filters}\label{s:receptive-simple-filters}

We now compute the receptive field geometry $(\alpha_h,\alpha_v,\beta_h,\beta_v,\Delta_h,\Delta_v)$ for the most common operators, namely filters. We consider in particular \emph{simple filters} that are characterised by an integer size, stride, and padding.

It suffices to reason in 1D.  Let $H'$ bet the vertical filter dimension, $S_h$ the subampling stride, and $P_h^-$ and $P_h^+$ the amount of zero padding applied to the top and the bottom of the input $\bx$. Here the value $y_{i''}$ depends on the samples:
\begin{align*}
 x_i : i
 &\in
 [1, H'] + S_h (i'' - 1) - P_h^-
=
\left[-\frac{H'-1}{2}, \frac{H'-1}{2}\right] + S_h (i''-1) - P_h^- + \frac{H'+1}{2}.
\end{align*}
Hence
\[
\alpha_h = S_h,
\qquad
\beta _h = \frac{H'+1}{2} - P_h^-,
\qquad
\Delta_h = H'.
\]
A similar relation holds for the horizontal direction.

Note that many blocks (e.g. max pooling, LNR, ReLU, most loss functions etc.) have a filter-like receptive field geometry. For example, ReLU can be considered a $1 \times 1$ filter, such that $H = S_h=1$ and $P_h^-=P_h^+ =0$. Note that in this case $\alpha_h=1$, $\beta_h=1$ and $\Delta_h=1$.

In addition to computing the receptive field geometry, we are often interested in determining the sizes of the arrays $\bx$ and $\by$ throughout the architecture. In the case of filters, and once more reasoning for a 1D slice, we notice that $y_i''$ can be obtained for $i''=1,2,\dots,H''$ where $H''$ is the largest value of $i''$ before the receptive fields falls outside $\bx$ (including padding). If $H$ is the height of the input array $\bx$, we get the condition
\[
   H' + S_h (H'' - 1) - P_h^- \leq H + P_h^+.
\]
Hence
\begin{equation}\label{e:filtered-height}
   H'' = \left\lfloor \frac{H - H' + P_h^- + P_h^+}{S_h} \right\rfloor + 1.	
\end{equation}

\subsection{Pooling in Caffe}

MatConvNet treats pooling operators like filters, using the rules above. In the library Caffe, this is done slightly differently, creating some incompatibilities. In their case, the pooling window is allowed to shift enough such that the last application always includes the last pixel of the input. If the stride is greater than one, this means that the last application of the pooling window can be partially outside the input boundaries even if padding is ``officially'' zero.

More formally, if $H'$ is the pool size and $H$ the size of the signal, the last application of the pooling window has index $i'' = H''$ such that
\[
  S_h(i''-1) + H' \big|_{i''= H''} \geq H
  \qquad
  \Leftrightarrow
  \qquad
  H'' = \left\lceil 
  \frac{H - H'}{S_h}
  \right\rceil
  + 1.
\]
If there is padding, the same logic applies after padding the input image, such that the output has height:
\[
H'' = \left\lceil 
  \frac{H - H' + P_h^- + P_h^+}{S_h}
  \right\rceil
  + 1.
\]
This is the same formula as for above filters, but with the ceil instead of floor operator. Note that in practice $P_h^- = P_h^+ = P_h$ since Caffe does not support asymmetric padding. 

Unfortunately, it gets more complicated. Using the formula above, it can happen that the last padding application is completely outside the input image and Caffe tries to avoid it. This requires
\begin{equation}\label{e:pooling-caffe-constr}
  S(i'' - 1) - P_h^- + 1 \big|_{i''= H''} \leq H
  \qquad
  \Leftrightarrow
  \qquad
  H'' \leq \frac{H - 1 + P_h^-}{S_h} + 1.	
\end{equation}

Using the fact that for integers $a,b$, one has $\lceil a/b \rceil = \lfloor (a+b-1)/b \rfloor$, we can rewrite the expression for $H''$ as follows
\begin{align*}
H'' = \left\lceil 
  \frac{H - H' + P_h^- + P_h^+}{S_h}
  \right\rceil
  + 1
  =
  \left\lfloor
  \frac{H - 1 +P_h^-}{S_h}
  +
  \frac{P^+_h + S_h - H'}{S_h}
  \right\rfloor
  +1.
 \end{align*}
Hence if $P_h^+ +  S_h \leq H' $ then the second term is less than zero and \eqref{e:pooling-caffe-constr} is satisfied. In practice, Caffe assumes that $P_h^+, P_h^- \leq H' -1$, as otherwise the first filter application falls entirely in the padded region.  Hence, we can upper bound the second term:
\[
\frac{P^+_h + S_h - H'}{S_h}
\leq
\frac{S_h - 1}{S_h}
\leq
1.
\]
We conclude that, for any choices of $P_h^+$ and $S_h$ allowed by Caffe, the formula above may violate constraint \eqref{e:pooling-caffe-constr} by at most one unit. Caffe has a special provision for that and lowers $H''$ by one when needed. Furthermore, we see that if $P_h^+=0$ \emph{and} $S_h \leq H'$ (which is often the case and may be assumed by Caffe), then the equation is also satisfied and Caffe skips the check.

Next, we find MatConvNet equivalents for these parameters. Assume that Caffe applies a symmetric padding $P_h$. Then in MatConvNet $P_h^-=P_h$ to align the top part of the output signal. To match Caffe, the last sample of the last filter application has to be on or to the right of the last Caffe-padded pixel:
\[
\underbrace{
S_h
\left(
\underbrace
{
\left\lfloor
\frac{H - H' + P_h^- + P_h^+}{S_h}  + 1 
\right\rfloor
}_{\text{MatConvNet rightmost pooling index}}
- 1
\right)
+ H'
}_{\text{MatConvNet rightmost pooled input sample}}
\geq
\underbrace{
H + 2P_h^-
}_{\text{Caffe rightmost input sample with padding}}.
\]
Rearranging
\[
\left\lfloor
\frac{H - H' + P_h^- + P_h^+}{S_h}
\right\rfloor
\geq
\frac{H - H' + 2P_h^{-}}{S_h}
\]
Using $\lfloor a/b \rfloor = \lceil (a - b + 1)/b\rceil$ we get the \emph{equivalent} condition:
\[
\left\lceil 
\frac{H - H' + 2P_h^-}{S_h} + \frac{P_h^+ - P_h^- - S_h + 1}{S_h}
\right\rceil
\geq
\frac{H - H' + 2P_h^-}{S_h} 
\]
Removing the ceil operator lower bounds the left-hand side of the equation and produces the \emph{sufficient} condition
\[
 P_h^+ \geq P_h^- + S_h - 1.
\]
As before, this may still be too much padding, causing the last pool window application to be entirely in the rightmost padded area. MatConvNet places the restriction $P_h^+ \leq H' -1$, so that
\[
  P_h^+ = \min\{ P_h^- + S_h - 1 , H' - 1\}.
\]
For example, a pooling region of width $H'=3$ samples with  a stride of $S_h=1$ samples and null Caffe padding $P_h^-=0$, would result in a right MatConvNet padding of $P_h^+ = 1$.

\section{Convolution transpose}\label{s:receptive-convolution-transpose}

The convolution transpose block is similar to a simple filter, but somewhat more complex. Recall that convolution transpose (\cref{s:impl-convolution-transpose}) is the transpose of the convolution operator, which in turn is a filter. Reasoning for a 1D slice, let $x_i$ be the input to the convolution transpose block and $y_{i''}$ its output. Furthermore let $U_h$, $C_h^-$, $C_h^+$ and $H'$ be the upsampling factor, top and bottom crops, and filter height, respectively.

If we look at the convolution transpose backward, from the output to the input (see also \cref{f:convt}), the data dependencies are the same as for the convolution operator, studied in \cref{s:receptive-simple-filters}. Hence there is an interaction between $x_i$ and $y_{i''}$ only if
\begin{equation}\label{e:convt-bounds}
   1 + U_h(i - 1) - C_h^- \leq i'' \leq H' + U_h(i - 1) - C_h^-
\end{equation}
where cropping becomes padding and upsampling becomes downsampling. Turning this relation around, we find that
\[
 \left\lceil \frac{i'' + C_h^- -H'}{S_h} \right\rceil + 1
 \leq
 i
 \leq
 \left\lfloor \frac{i'' + C_h^- - 1}{S_h} \right\rfloor + 1 .
\]
Note that, due to rounding, it is not possible to express this set tightly in the form outlined above. We can however relax these two relations (hence obtaining a slightly larger receptive field) and conclude that
\[
\alpha_h = \frac{1}{U_h},
\qquad
\beta_h = \frac{2C_h^- - H' + 1}{2 U_h} + 1,
\qquad
\Delta_h = \frac{H' -1}{U_h} + 1.
\]

Next, we want to determine the height $H''$ of the output $\by$ of convolution transpose as a function of the heigh $H$ of the input $\bx$ and the other parameters. Swapping input and output in  \eqref{e:filtered-height} results in the constraint:
\[
H = 1+ \left\lfloor \frac{H'' - H' + C_h^- + C_h^+}{U_h} \right\rfloor.
\]
If $H$ is now given as input, it is not possible to recover $H''$ uniquely from this expression; instead, all the following values are possible
\[
   S_h (H-1) +H' -  C_h^- - C_h^+ \leq H'' < S_h H +H' -  C_h^- - C_h^+.
\]
This is due to the fact that $U_h$ acts as a downsampling factor in the standard convolution direction and some of the samples to the right of the convolution input $\by$ may be ignored by the filter (see also \cref{f:conv} and \cref{f:convt}).

Since the height of $\by$ is then determined up to $S_h$ samples, and since the extra samples would be ignored by the computation and stay zero, we choose the tighter definition and set
\[
H'' =  U_h (H-1) +H' -  C_h^- - C_h^+.
\]

\section{Transposing receptive fields}\label{s:receptive-transposing}

Suppose we have determined that a later $\by = f(\bx)$ has a receptive field transformation $(\alpha_h,\beta_h,\Delta_h)$ (along one spatial slice). Now suppose we are given a block $\bx = g(\by)$ which is the ``transpose'' of $f$, just like the convolution transpose layer is the transpose of the convolution layer. By this, we mean that, if $y_{i''}$ depends on $x_{i}$ due to $f$, then $x_{i}$ depends on $y_{i''}$ due to $g$.

Note that, by definition of receptive fields, $f$ relates the  inputs and outputs index pairs $(i,i'')$ given by \eqref{e:receptive}, which can be rewritten as
\[
- \frac{\Delta_h-1}{2} \leq  i - \alpha_h (i'' -1) - \beta_h \leq\frac{\Delta_h-1}{2}.
\]
A simple manipulation of this expression results in the equivalent expression:
\[
- \frac{(\Delta_h + \alpha_h - 1)/\alpha_h-1}{2} \leq  i'' - \frac{1}{\alpha_h} (i - 1) - \frac{1 + \alpha_h - \beta_h }{\alpha_h} \leq\frac{(\Delta_h + \alpha_h - 1)/\alpha_h-1}{2\alpha_h}.
\]
Hence, in the reverse direction, this corresponds to a RF transformation
\[
\hat \alpha_h = \frac{1}{\alpha_h},
\qquad
\hat \beta_h = \frac{1 + \alpha_h - \beta_h}{\alpha_h},
\qquad
\hat \Delta_h = \frac{\Delta_h + \alpha_h -1}{\alpha_h}.
\]

\begin{example}
For convolution, we have found the parameters:
\[
\alpha_h = S_h,
\qquad
\beta_h = \frac{H'+1}{2} - P_h^-,
\qquad
\Delta_h = H'.
\]
Using the formulas just found, we can obtain the RF transformation for convolution transpose:
\begin{align*}
\hat \alpha_h &= \frac{1}{\alpha_h} = \frac{1}{S_h},
\\
\hat \beta_h &= \frac{1 + S_h - (H'+1)/2 + P_h^-}{S_h}
= \frac{P_h^- -H'/2 +1/2}{S_h} + 1
= \frac{2P_h^- -H' + 1}{S_h} + 1,
\\
\hat \Delta_h &= \frac{H' + S_h - 1}{S_h} = \frac{H' -1}{S_h} + 1.
\end{align*}
Hence we find again the formulas obtained in \cref{s:receptive-convolution-transpose}.
\end{example}

\section{Composing receptive fields}\label{s:receptive-composing}

Consider now the composition of two layers $h = g \circ f$ with receptive fields $(\alpha_f, \beta_f, \Delta_f)$ and $(\alpha_g, \beta_g, \Delta_g)$ (once again we consider only a 1D slice in the vertical direction, the horizontal one being the same). The goal is to compute the receptive field of $h$.

To do so, pick a sample $i_g$ in the domain of $g$. The first and last sample $i_f$ in the domain of $f$ to affect $i_g$ are given by:
\[
  i_f = \alpha_f (i_g- 1) + \beta_f \pm \frac{\Delta_f - 1}{2}.
\]
Likewise, the first and last sample $i_g$ to affect a given output sample $i_h$ are given by
\[
  i_g = \alpha_g (i_h- 1) + \beta_g \pm \frac{\Delta_g - 1}{2}.
\]
Substituting one relation into the other, we see that the first and last sample $i_f$ in the domain of $g \circ f$ to affect $i_h$ are:
\begin{align*}\
 i_f &= \alpha_f \left(\alpha_g (i_h- 1) + \beta_g \pm \frac{\Delta_g - 1}{2} - 1\right) + \beta_f \pm \frac{\Delta_f - 1}{2}	
 \\
&= \alpha_f\alpha_g (i_h-1)
 + \alpha_f \beta_g - 1 + \beta_f
 \pm \frac{\alpha_f (\Delta_g - 1) + \Delta_f -1}{2}.
\end{align*}
We conclude that
\[
\alpha_h = \alpha_f \alpha_g,
\qquad
\beta_h =  \alpha_f (\beta_g - 1) + \beta_f,
\qquad
\Delta_h = \alpha_f (\Delta_g - 1) + \Delta_f.
\]

\section{Overlaying receptive fields}\label{s:receptive-overlying}

Consider now the combination $h(f(\bx_1), g(\bx_2))$ where the domains of $f$ and $g$ are the same. Given the rule above, it is possible to compute how each output sample $i_h$ depends on each input sample $i_f$ through $f$ and on each input sample $i_g$ through $g$. Suppose that this gives receptive fields $(\alpha_{hf}, \beta_{hf}, \Delta_{hf})$ and $(\alpha_{hg}, \beta_{hg}, \Delta_{hg})$ respectively. Now assume that the domain of $f$ and $g$ coincide, i.e. $\bx = \bx_1 = \bx_2$. The goal is to determine the combined receptive field.

This is only possible if, and only if, $\alpha = \alpha_{hg} = \alpha_{hf}$. Only in this case, in fact, it is possible to find a sliding window receptive field that tightly encloses the receptive field due to $g$ and $f$ at all points according to formulas~\eqref{e:receptive}. We say that these two receptive fields are \emph{compatible}. The range of input samples $i = i_f = i_g$ that affect any output sample $i_h$ is then given by
\begin{align*}
	  i_\text{max}&=
  \alpha (i_h- 1) + a, & a = \min
  \left\{\beta_{hf}- \frac{\Delta_{hf} - 1}{2}, \beta_g - \frac{\Delta_{hg} - 1}{2}\right\},
  \\
  	  i_\text{min} &=
  \alpha (i_h- 1) + b, & b = \max
  \left\{\beta_{hf}+ \frac{\Delta_{hf} - 1}{2}, \beta_g + \frac{\Delta_{hg} - 1}{2}\right\}.
\end{align*}
We conclude that the combined receptive field is
\[
\alpha = \alpha_{hg} = \alpha_{hf},
\qquad
\beta = \frac{a+b}{2},
\qquad
\delta = b - a + 1.
\]

\chapter{Implementation details}\label{s:impl}

This chapter contains calculations and details.

\section{Convolution}\label{s:impl-convolution}

It is often convenient to express the convolution operation in matrix form. To this end, let $\phi(\bx)$ be the \verb!im2row! operator, extracting all $W' \times H'$ patches from the map $\bx$ and storing them as rows of a $(H''W'') \times (H'W'D)$ matrix. Formally, this operator is given by:
\[
   [\phi(\bx)]_{pq} \underset{(i,j,d)=t(p,q)}{=} x_{ijd}
\]
where the index mapping $(i,j,d) = t(p,q)$ is
\[
 i = i''+i'-1, \quad
 j = j''+j'-1, \quad
 p = i'' + H'' (j''-1), \quad
 q = i' + H'(j'-1) + H'W' (d-1).
\]
It is also useful to define the ``transposed'' operator \verb!row2im!:
\[
   [\phi^*(M)]_{ijd}
   =
   \sum_{(p,q) \in t^{-1}(i,j,d)}
   M_{pq}.
\]
Note that $\phi$ and $\phi^*$ are linear operators. Both can be expressed by a matrix $H\in\real^{(H''W''H'W'D) \times(HWD)}$ such that
\[
  \vv(\phi(\bx)) = H \vv(\bx), \qquad 
  \vv(\phi^*(M)) = H^\top \vv(M).
\]
Hence we obtain the following expression for the vectorized output (see~\cite{kinghorn96integrals}):
\[
 \vv\by = 
 \vv\left(\phi(\bx) F\right)
 =
 \begin{cases}
 (I \otimes \phi(\bx)) \vv F, & \text{or, equivalently,} \\
 (F^\top \otimes I) \vv \phi(\bx),
 \end{cases}
\]
where $F\in\mathbb{R}^{(H'W'D)\times K}$ is the matrix obtained by reshaping the array $\bff$ and $I$ is an identity matrix of suitable dimensions. This allows obtaining the following formulas for the derivatives:
\[
\frac{dz}{d(\vv F)^\top}
=
\frac{dz}{d(\vv\by)^\top}
(I \otimes \phi(\bx))
= \vv\left[ 
\phi(\bx)^\top 
\frac{dz}{dY}
\right]^\top
\]
where $Y\in\real^{(H''W'')\times K}$ is the matrix obtained by reshaping the array $\by$. Likewise:
\[
\frac{dz}{d(\vv \bx)^\top}
=
\frac{dz}{d(\vv\by)^\top}
(F^\top \otimes I)
\frac{d\vv \phi(\bx)}{d(\vv \bx)^\top}
=
\vv\left[ 
\frac{dz}{dY}
F^\top
\right]^\top
H
\]
In summary, after reshaping these terms we obtain the formulas:
\[
\boxed{
\vv\by = 
 \vv\left(\phi(\bx) F\right),
\qquad
\frac{dz}{dF}
=
\phi(\bx)^\top\frac{d z}{d Y},
\qquad
\frac{d z}{d X}
=
\phi^*\left(
\frac{d z}{d Y}F^\top
\right)
}
\]
where $X\in\real^{(H'W')\times D}$ is the matrix obtained by reshaping $\bx$. Notably, these expressions are used to implement the convolutional operator; while this may seem inefficient, it is instead a fast approach when the number of filters is large and it allows leveraging fast BLAS and GPU BLAS implementations.

\section{Convolution transpose}\label{s:impl-convolution-transpose}

In order to understand the definition of convolution transpose, let $\by$ to be obtained from $\bx$ by the convolution operator as defined in \cref{s:convolution} (including padding and downsampling).  Since this is a linear operation, it can be rewritten as $\vv \by = M \vv\bx$ for a suitable matrix $M$; convolution transpose computes instead $\vv \bx = M^\top \vv \by$.  While this is simple to describe in term of matrices, what happens in term of indexes is tricky. In order to derive a formula for the convolution transpose, start from standard convolution (for a 1D signal):
\[
   y_{i''} = \sum_{i'=1}^{H'} f_{i'} x_{S (i''-1) + i' - P_h^-}, 
   \quad
    1 \leq i'' \leq 1 + \left\lfloor \frac{H - H' + P_h^- + P_h^+}{S} \right\rfloor,
\]
where $S$ is the downsampling factor, $P_h^-$ and $P_h^+$ the padding, $H$ the length of the input signal, $\bx$ and $H'$ the length of the filter $\bff$. Due to padding, the index of the input data $\bx$ may exceed the range $[1,H]$; we implicitly assume that the signal is zero padded outside this range.

In order to derive an expression of the convolution transpose,  we make use of the identity $\vv \by^\top (M \vv \bx) = (\vv \by^\top M) \vv\bx = \vv\bx^\top (M^\top \vv\by)$. Expanding this in formulas:
\begin{align*}
\sum_{i''=1}^b y_{i''} 
\sum_{i'=1}^{W'} f_{i'} x_{S (i''-1) + i'  -P_h^-}
&=
\sum_{i''=-\infty}^{+\infty}
\sum_{i'=-\infty}^{+\infty} 
y_{i''}\ f_{i'}\ x_{S (i''-1) + i'  -P_h^-}
\\
&=
\sum_{i''=-\infty}^{+\infty}
\sum_{k=-\infty}^{+\infty} 
y_{i''}\ f_{k-S(i'' -1) + P_h^-}\ x_{k}
\\
&=
\sum_{i''=-\infty}^{+\infty}
\sum_{k=-\infty}^{+\infty} 
y_{i''}%
\ %
f_{%
(k-1+ P_h^-) \bmod S +
S \left(1 -i''  + \left\lfloor \frac{k-1+ P_h^-}{S} \right\rfloor\right)+1
}\ x_{k}
\\
&=
\sum_{k=-\infty}^{+\infty} 
x_{k}
\sum_{q=-\infty}^{+\infty}
y_{\left\lfloor \frac{k-1+ P_h^-}{S} \right\rfloor + 2 - q}
\ %
f_{(k-1+ P_h^-)\bmod S +S(q - 1)+1}.
\end{align*}
Summation ranges have been extended to infinity by assuming that all signals are zero padded as needed. In order to recover such ranges, note that $k \in [1,H]$ (since this is the range of elements of $\bx$ involved in the original convolution). Furthermore, $q\geq 1$ is the minimum value of $q$ for which the filter $\bff$ is non zero; likewise, $q\leq \lfloor (H'-1)/2\rfloor +1$ is a fairly tight upper bound on the maximum value (although, depending on $k$, there could be an element less). Hence
\begin{equation}\label{e:convt-step}
 x_k = 
 \sum_{q=1}^{1 + \lfloor \frac{H'-1}{S} \rfloor}
y_{\left\lfloor \frac{k-1+ P_h^-}{S} \right\rfloor + 2 - q}\ %
f_{(k-1+ P_h^-)\bmod S +S(q - 1)+1},
\qquad k=1,\dots, H.
\end{equation}
Note that the summation extrema in \eqref{e:convt-step} can be refined slightly to account for the finite size of $\by$ and $\bw$:
\begin{multline*}
\max\left\{
1, 
\left\lfloor \frac{k-1 + P_h^-}{S} \right\rfloor + 2 - H''
\right\}
\leq q \\
\leq
1 +\min\left\{
\left\lfloor \frac{H'-1-(k-1+ P_h^-)\bmod S}{S} \right\rfloor, 
\left\lfloor \frac{k-1 + P_h^-}{S} \right\rfloor
\right\}.
\end{multline*}
The size $H''$ of the output of convolution transpose is obtained in \cref{s:receptive-convolution-transpose}.

\section{Spatial pooling}\label{s:impl-pooling}

Since max pooling simply selects for each output element an input element, the relation can be expressed in matrix form as
$
    \vv\by = S(\bx) \vv \bx
$
for a suitable selector matrix $S(\bx)\in\{0,1\}^{(H''W''D) \times (HWD)}$. The derivatives can the be written as:
$
\frac{d z}{d (\vv \bx)^\top}
=
\frac{d z}{d (\vv \by)^\top}
S(\bx),
$
for all but a null set of points, where the operator is not differentiable (this usually does not pose problems in optimization by stochastic gradient). For max-pooling, similar relations exists with two differences: $S$ does not depend on the input $\bx$ and it is not binary, in order to account for the normalization factors. In summary, we have the expressions:
\begin{equation}\label{e:max-mat}
\boxed{
\vv\by = S(\bx) \vv \bx,
\qquad
\frac{d z}{d \vv \bx}
=
S(\bx)^\top
\frac{d z}{d \vv \by}.
}
\end{equation}

\section{Activation functions}\label{s:impl-activation}

\subsection{ReLU}\label{s:impl-relu}

The ReLU operator can be expressed in matrix notation as
\[
\vv\by = \diag\bfs \vv \bx,
\qquad
\frac{d z}{d \vv \bx}
=
\diag\bfs
\frac{d z}{d \vv \by}
\]
where $\bfs = [\vv \bx > 0] \in\{0,1\}^{HWD}$ is an indicator vector.

\subsection{Sigmoid}\label{s:impl-sigmoid}

The derivative of the sigmoid function is given by
\begin{align*}
\frac{dz}{dx_{ijk}}
&= 
\frac{dz}{d y_{ijd}} 
\frac{d y_{ijd}}{d x_{ijd}}
=
\frac{dz}{d y_{ijd}} 
\frac{-1}{(1+e^{-x_{ijd}})^2} ( - e^{-x_{ijd}})
\\
&=
\frac{dz}{d y_{ijd}} 
y_{ijd} (1 - y_{ijd}).
\end{align*}
In matrix notation:
\[
\frac{dz}{d\bx} = \frac{dz}{d\by} \odot 
\by \odot 
(\bone\bone^\top - \by).
\]

\section{Spatial bilinear resampling}\label{s:impl-sampler}

The projected derivative $d\langle \bp, \phi(\bx,\bg)\rangle / d\bx$ of the spatial bilinaer resampler operator with respect to the input image $\bx$ can be found as follows:
\begin{multline}\label{e:bilinear-back-x}
  \frac{\partial}{\partial x_{ijc}}
  \left[
  \sum_{i''j''c''}
  p_{i''k''c''}
  \sum_{i'=1}^H
  \sum_{j'=1}^W 
  x_{i'j'c''}
  \max\{0, 1-|\alpha_v g_{1i''j''} + \beta_v -i'|\}
  \max\{0, 1-|\alpha_u g_{2i''j''} + \beta_u -j'|\}
  \right]
  \\
=
  \sum_{i''j''}
  p_{i''k''c}
  \max\{0, 1-|\alpha_v g_{1i''j''} + \beta_v -i|\}
  \max\{0, 1-|\alpha_u g_{2i''j''} + \beta_u -j|\}.
\end{multline}
Note that the formula is similar to Eq.~\ref{e:bilinear}, with the difference that summation is on $i''$ rather than $i$.

The projected derivative $d\langle \bp, \phi(\bx,\bg)\rangle / d\bg$ with respect to the grid is similar:
\begin{multline}\label{e:bilinear-back-g}
  \frac{\partial}{\partial g_{1i'j'}}
  \left[
  \sum_{i''j''c}
  p_{i''k''c}
  \sum_{i=1}^H
  \sum_{j=1}^W 
  x_{ijc}
  \max\{0, 1-|\alpha_v g_{1i''j''} + \beta_v -i|\}
  \max\{0, 1-|\alpha_u g_{2i''j''} + \beta_u -j|\}
  \right]
  \\
=
  -
  \sum_c
  p_{i'j'c}
  \sum_{i=1}^H
  \sum_{j=1}^W
  \alpha_v x_{ijc}
  \max\{0, 1-|\alpha_v g_{2i'j'} + \beta_v -j|\}
  \sign(\alpha_v g_{1i'j'} + \beta_v -j)
  \mathbf{1}_{\{-1 < \alpha_u g_{2i'j'} + \beta_u < 1\}}.
\end{multline}
A similar expression holds for $\partial g_{2i'j'}$

\section{Normalization}\label{s:normalization}

\subsection{Local response normalization (LRN)}\label{s:impl-ccnormalization}

The derivative is easily computed as:
\[
\frac{dz}{d x_{ijd}}
=
\frac{dz}{d y_{ijd}}
L(i,j,d|\bx)^{-\beta}
-2\alpha\beta x_{ijd}
\sum_{k:d\in G(k)}
\frac{dz}{d y_{ijk}}
L(i,j,k|\bx)^{-\beta-1} x_{ijk} 
\]
where
\[
 L(i,j,k|\bx) = \kappa + \alpha \sum_{t\in G(k)} x_{ijt}^2.
\]

\subsection{Batch normalization}\label{s:impl-bnorm}

The derivative of the network output $z$ with respect to the multipliers $w_k$ and biases $b_k$ is given by
\begin{align*}
	\frac{dz}{dw_k} &= \sum_{i''j''k''t''}
\frac{dz}{d y_{i''j''k''t''}} 
\frac{d y_{i''j''k''t''}}{d w_k}
=
\sum_{i''j''t''}
\frac{dz}{d y_{i''j''kt''}} 
\frac{x_{i''j''kt''} - \mu_{k}}{\sqrt{\sigma_k^2 + \epsilon}},
\\
\frac{dz}{db_k} &= \sum_{i''j''k''t''}
\frac{dz}{d y_{i''j''k''t''}} 
\frac{d y_{i''j''k''t''}}{d w_k}
=
\sum_{i''j''t''}
\frac{dz}{d y_{i''j''kt''}}.
\end{align*}

The derivative of the network output $z$ with respect to the block input $x$ is computed as follows:
\[
\frac{dz}{dx_{ijkt}} = \sum_{i''j''k''t''}
\frac{dz}{d y_{i''j''k''t''}} 
\frac{d y_{i''j''k''t''}}{d x_{ijkt}}.
\]
Since feature channels are processed independently, all terms with $k''\not=k$ are zero. Hence
\[
\frac{dz}{dx_{ijkt}} = \sum_{i''j''t''}
\frac{dz}{d y_{i''j''kt''}} 
\frac{d y_{i''j''kt''}}{d x_{ijkt}},
\]
where
\[
\frac{d y_{i''j''kt''}}{d x_{ijkt}} 
=
w_k
\left(\delta_{i=i'',j=j'',t=t''} - \frac{d \mu_k}{d x_{ijkt}}\right)
\frac{1}{\sqrt{\sigma^2_k + \epsilon}}
-
\frac{w_k}{2}
\left(x_{i''j''kt''} - \mu_k\right)
\left(\sigma_k^2 + \epsilon \right)^{-\frac{3}{2}}
\frac{d \sigma_k^2}{d x_{ijkt}},
\]
the derivatives with respect to the mean and variance are computed as follows:
\begin{align*}
\frac{d \mu_k}{d x_{ijkt}} &= \frac{1}{HWT},
\\
\frac{d \sigma_k^2}{d x_{i'j'kt'}}
&=
\frac{2}{HWT}
\sum_{ijt}
\left(x_{ijkt} - \mu_k \right)
\left(\delta_{i=i',j=j',t=t'} - \frac{1}{HWT} \right)
=
\frac{2}{HWT} \left(x_{i'j'kt'} - \mu_k \right),
\end{align*}
and $\delta_E$ is the indicator function of the event $E$. Hence
\begin{align*}
\frac{dz}{dx_{ijkt}}
&=
\frac{w_k}{\sqrt{\sigma^2_k + \epsilon}}
\left(
\frac{dz}{d y_{ijkt}} 
-
\frac{1}{HWT}\sum_{i''j''kt''}
\frac{dz}{d y_{i''j''kt''}} 
\right)
\\
&-
\frac{w_k}{2(\sigma^2_k + \epsilon)^{\frac{3}{2}}}
\sum_{i''j''kt''}
\frac{dz}{d y_{i''j''kt''}} 
\left(x_{i''j''kt''} - \mu_k\right)
\frac{2}{HWT} \left(x_{ijkt} - \mu_k \right)
\end{align*}
i.e.
\begin{align*}
\frac{dz}{dx_{ijkt}}
&=
\frac{w_k}{\sqrt{\sigma^2_k + \epsilon}}
\left(
\frac{dz}{d y_{ijkt}} 
-
\frac{1}{HWT}\sum_{i''j''kt''}
\frac{dz}{d y_{i''j''kt''}} 
\right)
\\
&-
\frac{w_k}{\sqrt{\sigma^2_k + \epsilon}}
\,
\frac{x_{ijkt} - \mu_k}{\sqrt{\sigma^2_k + \epsilon}}
\,
\frac{1}{HWT}
\sum_{i''j''kt''}
\frac{dz}{d y_{i''j''kt''}} 
\frac{x_{i''j''kt''} - \mu_k}{\sqrt{\sigma^2_k + \epsilon}}.
\end{align*}
We can identify some of these terms with the ones computed as derivatives of bnorm with respect to $w_k$ and $\mu_k$:
\begin{align*}
\frac{dz}{dx_{ijkt}}
&=
\frac{w_k}{\sqrt{\sigma^2_k + \epsilon}}
\left(
\frac{dz}{d y_{ijkt}} 
-
\frac{1}{HWT}
\frac{dz}{d b_k} 
-
\frac{x_{ijkt} - \mu_k}{\sqrt{\sigma^2_k + \epsilon}}
\,
\frac{1}{HWT}
\frac{dz}{dw_k}
\right).
\end{align*}

\subsection{Spatial normalization}\label{s:impl-spnorm}

The neighbourhood norm $n^2_{i''j''d}$ can be computed by applying average pooling to $x_{ijd}^2$ using \verb!vl_nnpool! with a $W'\times H'$ pooling region, top padding $\lfloor \frac{H'-1}{2}\rfloor$, bottom padding $H'-\lfloor \frac{H-1}{2}\rfloor-1$, and similarly for the horizontal padding.

The derivative of spatial normalization can be obtained as follows:
\begin{align*}
\frac{dz}{dx_{ijd}} 
&= \sum_{i''j''d}
\frac{dz}{d y_{i''j''d}} 
\frac{d y_{i''j''d}}{d x_{ijd}}
\\
&=
\sum_{i''j''d}
\frac{dz}{d y_{i''j''d}} 
(1 + \alpha n_{i''j''d}^2)^{-\beta}
\frac{dx_{i''j''d}}{d x_{ijd}} 
-\alpha\beta
\frac{dz}{d y_{i''j''d}} 
(1 + \alpha n_{i''j''d}^2)^{-\beta-1}
x_{i''j''d}
\frac{dn_{i''j''d}^2}{d (x^2_{ijd})} 
\frac{dx^2_{ijd}}{d x_{ijd}}
\\
&=
\frac{dz}{d y_{ijd}} 
(1 + \alpha n_{ijd}^2)^{-\beta}
-2\alpha\beta x_{ijd}
\left[
\sum_{i''j''d}
\frac{dz}{d y_{i''j''d}} 
(1 + \alpha n_{i''j''d}^2)^{-\beta-1}
x_{i''j''d}
\frac{dn_{i''j''d}^2}{d (x_{ijd}^2)}
\right]
\\
&=
\frac{dz}{d y_{ijd}} 
(1 + \alpha n_{ijd}^2)^{-\beta}
-2\alpha\beta x_{ijd}
\left[
\sum_{i''j''d}
\eta_{i''j''d}
\frac{dn_{i''j''d}^2}{d (x_{ijd}^2)}
\right],
\quad
\eta_{i''j''d}=
\frac{dz}{d y_{i''j''d}} 
(1 + \alpha n_{i''j''d}^2)^{-\beta-1}
x_{i''j''d}
\end{align*}
Note that the summation can be computed as the derivative of the
\verb!vl_nnpool! block.

\subsection{Softmax}\label{s:impl-softmax}

Care must be taken in evaluating the exponential in order to avoid underflow or overflow. The simplest way to do so is to divide the numerator and denominator by the exponential of the maximum value:
\[
 y_{ijk} = \frac{e^{x_{ijk} - \max_d x_{ijd}}}{\sum_{t=1}^D e^{x_{ijt}- \max_d x_{ijd}}}.
\]
The derivative is given by:
\[
\frac{dz}{d x_{ijd}}
=
\sum_{k}
\frac{dz}{d y_{ijk}}
\left(
e^{x_{ijd}} L(\bx)^{-1} \delta_{\{k=d\}}
-
e^{x_{ijd}}
e^{x_{ijk}} L(\bx)^{-2}
\right),
\quad
L(\bx) = \sum_{t=1}^D e^{x_{ijt}}.
\]
Simplifying:
\[
\frac{dz}{d x_{ijd}}
=
y_{ijd} 
\left(
\frac{dz}{d y_{ijd}}
-
\sum_{k=1}^K
\frac{dz}{d y_{ijk}} y_{ijk}.
\right).
\]
In matrix form:
\[
  \frac{dz}{dX} = Y \odot \left(\frac{dz}{dY} 
  - \left(\frac{dz}{dY} \odot Y\right) \bone\bone^\top\right)
\]
where $X,Y\in\real^{HW\times D}$ are the matrices obtained by reshaping the arrays
$\bx$ and $\by$. Note that the numerical implementation of this expression is straightforward once the output $Y$ has been computed with the caveats above.

\section{Categorical losses}\label{s:impl-losses}

This section obtains the projected derivatives of the categorical losses in \cref{s:losses}. Recall that all losses give a scalar output, so the projection tensor $p$ is trivial (a scalar).

\subsection{Classification losses}\label{s:impl-loss-classification}

\paragraph{Top-$K$ classification error.} The derivative is zero a.e.\

\paragraph{Log-loss.} The projected derivative is:
\[
\frac{\partial p \ell(\bx,c)}{\partial x_k}
=
- p \frac{\partial \log (x_c) }{\partial x_k}
=
- p x_c \delta_{k=c}.
\]

\paragraph{Softmax log-loss.} The projected derivative is given by:
\[
\frac{\partial p \ell(\bx,c)}{\partial x_k}
=
- p \frac{\partial}{\partial x_k}
\left(x_c - \log \sum_{t=1}^C e^{x_t}\right)
=
- p \left(\delta_{k=c} - \frac{e^{x_c}}{\sum_{t=1}^C e^{x_t}} \right).
\]
In brackets, we can recognize the output of the loss itself:
\[
 y = \ell(\bx,c) = \frac{e^{x_c}}{\sum_{t=1}^C e^{x_t}}.
\]
Hence the loss derivatives rewrites:
\[
\frac{\partial p \ell(\bx,c)}{\partial x_k}
=
- p \left(\delta_{k=c} - y \right).
\]

\paragraph{Multi-class hinge loss.} The projected derivative is:
\[
\frac{\partial p \ell(\bx,c)}{\partial x_k}
=
- p\,\mathbf{1}[x_c < 1]\,\delta_{k=c}.
\]

\paragraph{Structured multi-class hinge loss.} The projected derivative is:
\[
\frac{\partial p \ell(\bx,c)}{\partial x_k}
=
- p\,\mathbf{1}[x_c < 1 + \max_{t\not= c} x_t]\,(\delta_{k=c} - \delta_{k=t^*}),
\qquad
t^* = \argmax_{t =1,2,\dots,C} x_t.
\]

\subsection{Attribute losses}\label{s:impl-loss-attribute}

\paragraph{Binary error.} The derivative of the binary error is 0 a.e.

\paragraph{Binary log-loss.} The projected derivative is:
\[
\frac{\partial p \ell(x,c)}{\partial x}
=
- p \frac{c}{c \left(x - \frac{1}{2}\right) + \frac{1}{2}}.
\]

\paragraph{Binary logistic loss.} The projected derivative is:
\[
\frac{\partial p \ell(x,c)}{\partial x}
=
- p \frac{\partial}{\partial x} \log \frac{1}{1+e^{-cx}}
=
- p \frac{c e^{-cx}}{1 + e^{-cx}}
=
- p \frac{c}{e^{cx} + 1}
=
- pc\, \sigma(-cx).
\]

\paragraph{Binary hinge loss.} The projected derivative is
\[
\frac{\partial p \ell(x,c)}{\partial x}
=
- pc\,\mathbf{1}[cx < 1].
\]

\section{Comparisons}\label{s:impl-comparisons}

\subsection{$p$-distance}\label{s:impl-pdistance}

The derivative of the operator without root is given by:
\begin{align*}
\frac{dz}{dx_{ijd}}
&=
\frac{dz}{dy_{ij}}
p |x_{ijd} - \bar x_{ijd}|^{p-1} \operatorname{sign} (x_{ijd} - \bar x_{ijd}).
\end{align*}
The derivative of the operator with root is given by:
\begin{align*}
\frac{dz}{dx_{ijd}}
&=
\frac{dz}{dy_{ij}}
\frac{1}{p}
\left(\sum_{d'} |x_{ijd'} - \bar x_{ijd'}|^p \right)^{\frac{1}{p}-1}
p |x_{ijd} - \bar x_{ijd}|^{p-1} \sign(x_{ijd} - \bar x_{ijd})
\\
&= 
\frac{dz}{dy_{ij}}
\frac{|x_{ijd} - \bar x_{ijd}|^{p-1} \sign(x_{ijd} - \bar x_{ijd})}{y_{ij}^{p-1}};
\frac{dz}{d\bar x_{ijd}} = -\frac{dz}{dx_{ijd}}.
\end{align*}
The formulas simplify a little for $p=1,2$ which are therefore implemented as special cases.

\clearpage

\bibliographystyle{plain}
\bibliography{references,/Users/vedaldi/src/bibliography/vedaldi}

\begin{thebibliography}{10}

\bibitem{chatfield14return}
K.~Chatfield, K.~Simonyan, A.~Vedaldi, and A.~Zisserman.
\newblock Return of the devil in the details: Delving deep into convolutional
  nets.
\newblock In {\em Proc. {BMVC}}, 2014.

\bibitem{deng09imagenet}
J.~Deng, W.~Dong, R.~Socher, L.-J. Li, K.~Li, and L.~Fei-Fei.
\newblock {ImageNet: A Large-Scale Hierarchical Image Database}.
\newblock In {\em Proc. {CVPR}}, 2009.

\bibitem{ioffe15batch}
S.~Ioffe and C.~Szegedy.
\newblock Batch normalization: Accelerating deep network training by reducing
  internal covariate shift.
\newblock {\em CoRR}, 2015.

\bibitem{ioffe2015}
S.~{Ioffe} and C.~{Szegedy}.
\newblock {Batch Normalization: Accelerating Deep Network Training by Reducing
  Internal Covariate Shift}.
\newblock {\em ArXiv e-prints}, 2015.

\bibitem{jia13caffe}
Yangqing Jia.
\newblock {Caffe}: An open source convolutional architecture for fast feature
  embedding.
\newblock \url{http://caffe.berkeleyvision.org/}, 2013.

\bibitem{kinghorn96integrals}
D.~B. Kinghorn.
\newblock Integrals and derivatives for correlated gaussian fuctions using
  matrix differential calculus.
\newblock {\em International Journal of Quantum Chemestry}, 57:141--155, 1996.

\bibitem{krizhevsky12imagenet}
A.~Krizhevsky, I.~Sutskever, and G.~E. Hinton.
\newblock Imagenet classification with deep convolutional neural networks.
\newblock In {\em Proc. {NIPS}}, 2012.

\bibitem{lin13network}
Min Lin, Qiang Chen, and Shuicheng Yan.
\newblock Network in network.
\newblock {\em CoRR}, abs/1312.4400, 2013.

\bibitem{simonyan14deep}
K.~Simonyan, A.~Vedaldi, and A.~Zisserman.
\newblock Deep inside convolutional networks: Visualising image classification
  models and saliency maps.
\newblock In {\em Proc. {ICLR}}, 2014.

\bibitem{simonyan15very}
K.~Simonyan and A.~Zisserman.
\newblock Very deep convolutional networks for large-scale image recognition.
\newblock 2015.

\bibitem{vedaldi10vlfeat}
A.~Vedaldi and B.~Fulkerson.
\newblock {VLFeat} -- {An} open and portable library of computer vision
  algorithms.
\newblock In {\em Proc. {ACM} Int. Conf. on Multimedia}, 2010.

\bibitem{zeiler14visualizing}
M.~D. Zeiler and R.~Fergus.
\newblock Visualizing and understanding convolutional networks.
\newblock In {\em Proc. {ECCV}}, 2014.

\end{thebibliography}
\end{document}